\documentclass[journal]{IEEEtran}

\usepackage{cite}

\usepackage{amsmath}
\usepackage{mathrsfs}

\usepackage{algorithmic}

\usepackage{array}

\usepackage{subfigure}

\usepackage{url}
\usepackage{graphicx,amsmath,amssymb,amsfonts}
\usepackage{algorithmic,algorithm}

\usepackage{setspace}

\usepackage{xcolor}

\usepackage{epstopdf}

\newtheorem{theorem}{\textbf{Theorem}}

\newtheorem{lemma}{\textbf{Lemma}}

\newtheorem{corollary}{\textbf{Corollary}}

\makeatletter

\newcommand{\Rmnum}[1]{\expandafter\@slowromancap\romannumeral #1@}
\makeatother

\newcommand{\tabincell}[2]{\begin{tabular}{@{}#1@{}}#2\end{tabular}}

\usepackage{framed}

\usepackage{cancel}

\begin{document}
	\bstctlcite{ref:BSTcontrol}
	
	\title{Balancing Accuracy and Integrity for Reconfigurable Intelligent Surface-aided Over-the-Air Federated Learning}
	
	\author{Jingheng~Zheng,
				 Hui~Tian, \IEEEmembership{Senior~Member,~IEEE,}
				 Wanli~Ni, \IEEEmembership{Graduate~Student~Member,~IEEE,}
				 Wei~Ni,~\IEEEmembership{Senior~Member,~IEEE,}
				 and~Ping~Zhang,~\IEEEmembership{Fellow,~IEEE}
		\thanks{This work was funded by Beijing University of Posts and Telecommunications-China Mobile Reserch Institute Joint Innovation Center. This paper has been published in part at the IEEE International Symposium on Personal, Indoor and Mobile Radio Communications (PIMRC), Virtual, September 2021, DOI:~10.1109/PIMRC50174.2021.9569612.\nocite{Zheng2021QoS} \textit{(Corresponding author: Hui Tian.)}}
		\thanks{J.~Zheng,~H.~Tian,~W.~Ni,~and~P.~Zhang are with the State Key Laboratory of Networking and Switching Technology, Beijing University of Posts and Telecommunications, Beijing 100876, China (e-mail: zhengjh@bupt.edu.cn;~tianhui@bupt.edu.cn;~charleswall@bupt.edu.cn;~pzhang@bupt.edu.cn).}
		\thanks{W.~Ni is with the Commonwealth Scientific and Industrial Research Organization (CSIRO), Sydney NWS 2122, Australia (e-mail: wei.ni@data61.csiro.au).}
		
		\vspace{-0.4 cm}
	}
	
	\maketitle

	\begin{abstract}
		Over-the-air federated learning (AirFL) allows devices to train a learning model in parallel and synchronize their local models using over-the-air computation. 
		The integrity of AirFL is vulnerable due to the obscurity of the local models aggregated over-the-air. 
		This paper presents a novel framework to balance the accuracy and integrity of AirFL, where multi-antenna devices and base station (BS) are jointly optimized with a reconfigurable intelligent surface (RIS). 
		The key contributions include a new  and non-trivial problem jointly considering the model accuracy and integrity of AirFL, and a new framework that transforms the problem into tractable subproblems. 
		Under perfect channel state information (CSI), the new framework minimizes the aggregated model's distortion and retains the local models' recoverability by optimizing the transmit beamformers of the devices, the receive beamformers of the BS, and the RIS configuration in an alternating manner.
		Under imperfect CSI, the new framework delivers a robust design of the beamformers and RIS configuration to combat non-negligible channel estimation errors.
		As corroborated experimentally, the novel framework can achieve comparable accuracy to the ideal FL while preserving local model recoverability under perfect CSI, and improve the accuracy when the number of receive antennas is small or moderate under imperfect CSI.
	\end{abstract}
	
	
	\section{Introduction}
	\IEEEPARstart{A}{s} a promising distributed machine learning (ML) framework, federated learning (FL) allows multiple workers to train a model in parallel based on their local datasets, thereby protecting the data privacy of the workers and accelerating the training~~\cite{B.2017Federated,Chen2021Convergence,Brendan2016Communication}.
	FL requires locally trained models to be aggregated periodically, to create the global model~\cite{Li2020Federated,Khan2021Federated}.
	Incorporating over-the-air computation (AirComp)~\cite{Zhu2019MIMO} into FL, over-the-air FL (AirFL) provides an efficient means to aggregate local models. 
	It allows the workers to upload their models using the same time-frequency resources, and obtain nomographic functions of the ML models directly by exploiting the superposition property of radio~\cite{Goldenbaum2015Nomographic,Yang2020Federated,Li2019Wirelessly}. 
	AirFL is suitable for wireless networks, where many distributed devices act as workers and their serving base station (BS) is the model aggregator.

	Reconfigurable intelligent surface (RIS) is an increasingly widely accepted technology, and is envisaged to be one of the promising enhancements for future wireless systems~\cite{Liu2021Reconfigurable}.
	The consideration of RISs is indispensable for a future-proof design of AirFL systems.
	The deployment of an RIS ushers in a new degree of freedom to augment the radio propagation environment (in addition to the transmit and receive beamforming). The RIS can be configured to alleviate the distortion of the aggregated model by tuning the phase shifts of its reflecting elements~\cite{Wu2019Intelligent}.
	Compared to traditional multiple-input-multiple-output (MIMO) AirComp systems, e.g.,~\cite{Wen2019Reduced}, the incorporation of an RIS confronts not only a new challenge of a different problem formulation with many more variables, but the unit-modulus constraints of the new variables and their coupling increase the complexity of the problem dramatically.
	
	A general challenge arising from general AirFL systems, including those with or without RIS, is the integrity of AirFL, as studied in our paper. The model integrity accounts collectively for the trustworthiness of the local models provided to the model aggregator, i.e., the BS, to produce the global model~\cite{Wahab2021Federated} and the accountability of the devices that produce the local models~\cite{Wang2016Identity}. While enjoying the substantially reduced requirement of radio resources and thus enhanced scalability, AirFL obscures the local models at the BS and prevents the BS from assessing the trustworthiness of the local models. This makes AirFL vulnerable to model poisoning attacks.
	Consider multi-antenna devices and BS, and an RIS comprising a large number of reconfigurable phase shifts. The optimization variables include the transmit beamformers of the devices, the receive beamformer of the BS, and the phase shifts of the RIS, and typically coupled. The optimization is generally non-convex and mathematically intractable, even when the perfect channel state information (CSI) is available~\cite{Ni2021FL}. 
	Leave alone the typically imperfect CSI in practice~\cite{Zhao2021Exploiting,Zeng2020joint}. 
	No existing study has considered the trustworthiness of the local models and the accountability of the devices producing the local models.
	
	\subsection{Related Work}
	The accuracy of FL systems has been used as the sole goal in most of the existing literature.
	The authors of~\cite{Zhu2020Broadband} developed a broadband analog aggregation scheme to aggregate the concurrently transmitted local model updates over the air. Two trade-offs between the signal-to-noise ratio (SNR) and truncation, and between reliability and quantity, were revealed. Compared to conventional orthogonal transmissions, the communication latency was significantly reduced.
	The authors of~\cite{Zhu2021One} proposed one-bit broadband digital aggregation to overcome the difficult deployment of analog modulation required by over-the-air aggregation, where the devices apply one-bit quantization to the stochastic gradient and the BS employs a majority-vote based decoder to estimate the aggregated gradient. Convergence analysis was carried out separately under channel noise, fading, and estimation errors.
	The authors of~\cite{Cao2022Transmission} studied the transmit power control of AirFL to reduce the aggregation errors. A closed-form optimality gap was derived to capture the impact of aggregation errors on the convergence behavior. The training latency was minimized against a given optimality gap.

	The authors of~\cite{Wen2019Reduced} designed MIMO AirComp to achieve fast wireless data aggregation for sensors of different clusters. Aiming to minimize the MSE of the received and aggregated signals, closed-form aggregate beamforming at the BS was designed by exploiting the low rank characteristics of the clustered channels. Two low-latency simultaneous channel feedback schemes were developed to retrieve a function of individual CSI in both disjoint and overlapping clusters.
	The authors of~\cite{Zhu2019MIMO} utilized AirComp to achieve efficient wireless data aggregation.
	The beamforming matrices of multi-antenna devices and a multi-antenna BS were optimized by applying a differential geometry technique to minimize the mean square error (MSE) of the received signals.
	The authors of~\cite{Yang2020Federated} extended AirComp to FL systems for fast model aggregation, and maximized supportable devices by optimizing the receive beamforming vector with difference-of-convex-functions (DC) programming.
	The authors of~\cite{Mohammad2020Machine} investigated both digital and analog FL schemes. 
	In the case of analog FL, the local gradients were first sparsified and projected to a lower-dimensional space, and then aggregated over the air.

	Incorporating the RIS into AirFL systems, the authors of~\cite{Ni2021StarRIS} jointly optimized the configuration of an RIS and the power allocation of devices to promote the convergence of AirFL.
	The authors of~\cite{Ni2021FL} aimed to improve the learning accuracy of an AirFL system comprising a single-antenna BS, multiple single-antenna devices and multiple RISs.
	The selection and power allocation of the devices, the receive amplification of the BS, and the phase shifts of the RISs were jointly optimized to minimize the MSE and select as many devices as possible.
	A non-convex bi-criterion problem was formulated and solved using alternating optimization (AO).
	The MSE was minimized using semidefinite relaxation (SDR) and successive convex approximation (SCA).
	The devices were selected using DC programming.
	The study did not consider the integrity of AirFL.
	As a matter of fact, no existing studies have considered the integrity of AirFL.

	Some recent studies have proposed algorithms and protocols to deliver the integrity of conventional FL, typically under the assumption of error-free channels. 
	The authors of~\cite{Blanchard2017Machine} investigated the impact of Byzantine attacks on FL.
	An algorithm, named Krum, was proposed to preclude Byzantine workers by selecting the worker with the minimum sum squared distance of its ML model.
	The authors of~\cite{So2021Byzantine} extended the Krum algorithm to improve the resilience to Byzantine attacks by selecting the most plausible set of users for model aggregation.
	The authors of~\cite{Xu2020VerifyNet} proposed a VerifyNet framework to ensure the confidentiality of local models by designing a double-masking protocol, and verified the correctness of the aggregated model by using a homomorphic hash function.
	However, these works are inapplicable to AirFL, since they relied on the recoverability of the local models at the model aggregators.

	In a different yet relevant context, robust designs of beamformers and RIS configurations have been studied for communication systems under imperfect CSI.
	The authors of~\cite{Ang2019Robust} studied the robust design of beamforming vectors to minimize the MSE under the expected channel and the worst-case channel, where over-the-air signaling was used to generate nomographic functions between the workers and the aggregator.
	Considering an RIS with imperfect CSI, the authors of~\cite{Zhao2021Exploiting} and~\cite{Zeng2020joint} conducted a robust design of phase shifts and beamforming matrices in the downlink and uplink of a multi-user MIMO system, respectively.
	In~\cite{Liu2021Intelligent}, the signal-to-interference-and-noise ratio (SINR) was modeled based on historical SINRs and instantaneous CSI estimates.
	The phase shifts of an RIS and the power allocation of the devices were optimized to minimize the total transmit power using block-coordinate descent.
	These robust designs cannot directly apply to AirFL, because of distinct problems and settings. 
	
	\vspace{-0.2 cm}
	\subsection{Contribution and Organization}
	\label{contributions_and_organization}
	This paper presents a novel framework, which strikes a balance between accuracy and integrity for an AirFL system comprising multi-antenna devices and BS, and an RIS. 
	The key idea is that we propose to recover the local models serially using successive interference cancellation (SIC).
	The BS dedicates one receive beamformer for model aggregation, and the other receive beamformer for recovering local models.
	The two receive beamformers are jointly optimized with the transmit beamformers of the devices and the phase shifts of the RIS, to minimize the MSE of the model aggregation while maintaining sufficient power gaps between the local models for successful recovery.
	Another important aspect is that we develop new iterative algorithms which decompose this non-convex joint optimization problem into tractable subproblems. AO is employed to orchestrate the DC and SCA methods for optimizing the beamformers and phase shifts, first under perfect CSI and then imperfect CSI.

	The contributions of this paper are summarized as follows:
	\begin{enumerate}
		\item A novel system is proposed to balance the accuracy and integrity of RIS-aided AirFL. The BS dedicates two receive beamformers separately for AirFL model aggregation and SIC-based local model recovery. To the best of our knowledge, no existing study has considered the integrity of AirFL. Let alone RIS-aided AirFL (of which AirFL is a special case).
		\item A new problem is formulated to minimize the MSE of the model aggregation, subject to the sufficient power gaps between the local models for effective model recovery. Both perfect and imperfect CSI are considered between the devices, BS and RIS.
		\item A new AO-based algorithm is developed to solve the problem under perfect CSI by optimizing the receive beamformers of the BS, the transmit beamformers of the devices, and the phase shifts of the RIS in an alternating manner.
		\item Non-trivial efforts are devoted to convexifying the optimizations using DC programming and SCA. An analytic expression is derived for the second-order coefficient of the Taylor expansion adopted to approximate the surrogate functions of the SCA-based phase shift configuration, substantially reducing the complexity compared to the standard Armijo rule-based iterative search for the coefficient.
		\item The new algorithm is extended under imperfect CSI, and showcases its viability and robustness in the presence of non-negligible channel estimation errors.
	\end{enumerate}

	The new framework is experimentally evaluated based on MNIST/Fashion-MNIST dataset using a multilayer perceptron (MLP).
	Under perfect CSI, the framework can achieve comparable learning accuracy to the ideal FL, and retain the recoverability of the local models.
	The RIS may increase the susceptibility of AirFL to imperfect CSI when the transmit power is higher or there are a large number of receive antennas.
	Nevertheless, the RIS can improve the accuracy under imperfect CSI, when the number of receive antennas is small or moderate.
	
	The remainder of this paper is organized as follows.
	The system architecture is presented in Section~\ref{system_model}.
	The problem formulation and the proposed beamformer design and RIS configuration are developed under perfect CSI in Section~\ref{solution_perfect_CSI}, followed by a robust design under imperfect CSI in Section~\ref{solution_imperfect_CSI}.
	Section~\ref{numerical_results} provides experimental results, followed by conclusions in Section~\ref{conclusion}.
	
	\emph{Notations}: Lower- and upper-case boldface indicate vector and matrix,  respectively;
	${\bf{I}}_{N}$ denotes the $N \times N$ identity matrix;
	${{\bf{0}}_{M \times N}}$ denotes the $M \times N$ all-zero matrix;
	$\|{\bf{\cdot}}\|_{2}$ and $\|{\bf{\cdot}}\|_{F}$ denote matrix $2$-norm and Frobenius norm, respectively;
	${(\cdot)}^{\rm H}$, ${(\cdot)}^{\rm T}$, ${(\cdot)}^{-1}$ and $\text{tr}({\cdot})$ denote conjugate transpose, transpose, inverse and trace, respectively;
	$|\cdot|$ and ${\mathop{\rm Re }\nolimits} \{ \cdot \}$ denote the modulus and real part of a complex value, respectively;
	$\|\cdot\|$ and ${\rm diag}(\cdot)$ denote vector $2$-norm and diagonal matrix;
	$\langle \cdot, \cdot \rangle$ takes inner product;
	$\mathbb{E}[\cdot]$ takes statistical expectation;
	$\otimes$ and $\circ$ denote the Kronecker and Hadamard products, respectively;
	$\mathbb{C}^{{M} \times {N}}$ is the set of $M \times N$ complex matrices; 
	and $\mathbb{C}$ and $\mathbb{R}$ are the sets of complex and real numbers, respectively.
	
	\section{System Overview} \label{system_model}
	\begin{figure}[t]
		\centering
		\includegraphics[scale=0.55]{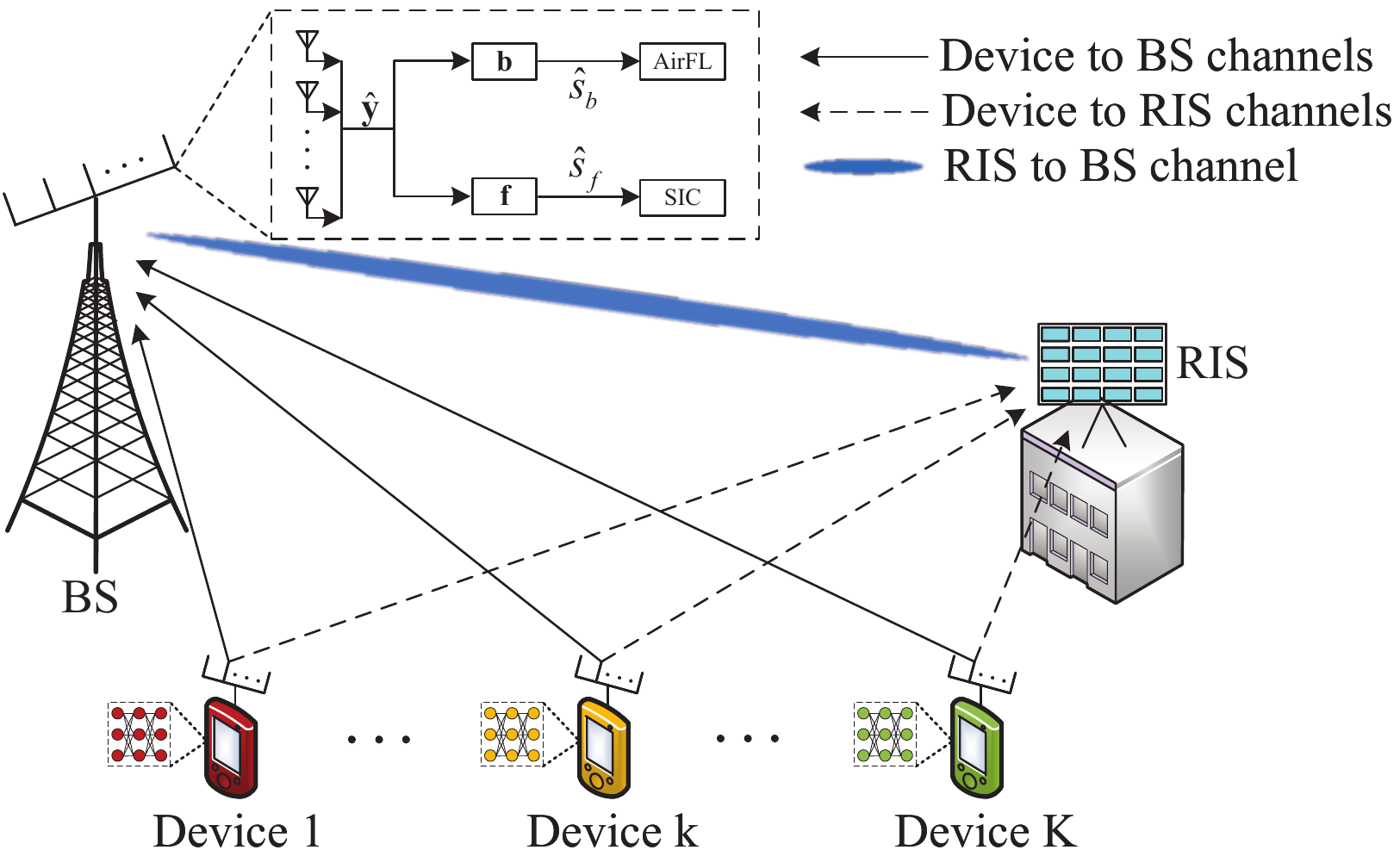}
		\vspace{-0.2 cm}
		\caption{An illustration on an RIS-aided MIMO AirFL system, where the BS conducts the model aggregation of AirFL and the serial recovery of the local models.}
		\label{system_model_figure}
		\vspace{-0.6 cm}
	\end{figure}
	
	As shown in Fig.~\ref{system_model_figure}, we consider an RIS-aided AirFL system, where there is a multi-antenna BS, $K$ multi-antenna devices, and an RIS.
	Each device has ${N_t}$ transmit antennas.
	The BS has ${N_r}$ receive antennas.
	%
	%
	The RIS has $M$ reflecting elements.
	$\mathcal{K} \buildrel \Delta \over =  \{ 1,2,\cdots,K\}$ collects the indexes to the devices.
	$\mathcal{M} \buildrel \Delta \over = \{ 1,2,\cdots,M\}$ collects the indexes to the reflecting elements of the RIS. 
	The phase shift of the $m$-th reflecting element, denoted by ${\phi _m}, m \in \mathcal{M}$, is within the range of $\left[ {0,2\pi } \right)$.
	${\bf{\Theta }} \buildrel \Delta \over = {\rm diag}( {e^{j{\phi _1}}},{e^{j{\phi _2}}},\cdots,{e^{j{\phi _M}}} )$ is the phase shift matrix of the RIS.

	The devices employ the mini-batch gradient descent method to train their local models $\{{\bf{w}}_k \} $ with their private datasets, and upload the models to the BS using the same time-frequency resources.
	We assume that all devices are synchronized\footnote{The authors of~\cite{Abari2015AirShare} designed the BS to broadcast a shared clock to all devices before their concurrent transmissions. To avoid the frequency offset among the devices, the BS also sends two single tones with their frequency difference matching the shared clock. By this means, all devices can be synchronized in both the time and frequency domains.}, as in~\cite{Cao2022Transmission,Mohammad2020Machine,Zhu2020Broadband,Mohammad2021Blind,Abari2015AirShare}.
	Using AirFL, the BS aggregates the local models $\{{\bf{w}}_k\}$ by computing the nomographic function and produces the global model ${\bf{w}}$. 
	In each communication round, the aggregation is written as
	\begin{equation}
		{\bf{w}} = \psi \left( {\sum\nolimits_{k = 1}^K {{\varphi _k}\left( {{{\bf{w}}_k}} \right)} } \right),
	\end{equation}
	where ${\varphi _k}(\cdot)$ and $\psi(\cdot)$ are the pre-processing function at the $k$-th device and the post-processing function at the BS, respectively.
	
	Consider the $k$-th device. The local model ${\bf{w}}_k$ is transformed to a sequence of transmit symbols arranged in a complex vector ${\bf{s}}_k$ by the pre-processing function ${\varphi _k}\left( \cdot \right)$, i.e., ${\bf{s}}_k={\varphi _k}\left( {{{\bf{w}}_k}} \right)$. The $k$-th device transmits the elements of the vector ${\bf{s}}_k$, denoted by a complex scalar ${s_k} \in \mathbb{C}$, sequentially to the BS, one element after another, in a communication round.
	At the BS, the desired superposition signal is $s = \sum\nolimits_{k = 1}^K {{s_k}}$.
	This framework can support dropout techniques typically used to reduce the size of the local models to be uploaded. For example, a federated dropout scheme was developed in~\cite{Wen2022Federated} to prune a global model into multiple subsets with different dropout rates adapting to the different abilities of the devices. The federated dropout can be executed at the devices to prune their models before the pre-processing.
	
	Consider a block fading channel. 
	The channel fading remains unchanged within a communication round of AirFL (i.e., a block) and changes independently between communication rounds~\cite{Cao2022Transmission,Xu2021Learning}. The duration of a communication round depends on the coherence time of the channel.
	Within a communication round, ${ {\bf{H}}_{d,k}} \in \mathbb{C}^{{N_r} \times {N_t}}$ denotes the channel matrix of the direct path from the $k$-th device to the BS; ${ {\bf{H}}_{r,k}} \! \in \! \mathbb{C}^{M \times {N_t}}$ denotes the channel matrix from the $k$-th device to the RIS; and ${ {\bf{G}}} \in \mathbb{C}^{M \times {N_r}}$ denotes the channel matrix from the RIS to the BS~\cite{Zhao2021Exploiting}.

	To estimate the channels, a transmitter sends full-rank pilot signals via its transmit antennas, as considered in~\cite{Zhou2021Joint,de2021Channel}. 
	The pilot signals are repeated $L$ times and each time the RIS is reconfigured. 
	Each of the RIS configurations, i.e., the phase shift matrix, is full-rank.
	In~\cite{Zhou2021Joint}, the received pilot signals were reorganized in the form of a multi-path signal. 
	Each of the paths corresponds to one of the RIS elements or the LoS path. 
	The minimum MSE (MMSE) method was taken to estimate the LoS path.
	In~\cite{de2021Channel}, the received pilot signals were arranged in a tensor, showing that the cascaded channel from the transmitter to the RIS and then the receiver exhibits the Khatri-Rao structure in the absence of the LoS path. 
	The individual channels between the RIS and the transmitter/receiver, i.e., $\bf{G}$ and ${\bf{H}}_{r,k}$, were estimated using the Khatri-Rao factorization algorithm.
	One can potentially run the algorithms developed in~\cite{Zhou2021Joint} and~\cite{de2021Channel} sequentially to first estimate the LoS path, and then cancel it to estimate the RIS-reflected channels. 
		
	Suppose that all $K$ devices send their pilot signals, one after another, to allow the BS to estimate the channels. 
	The signaling overhead is ${\tau}LK$ (symbols), where $\tau$ is the number of symbols in a pilot signal.
	It is also possible to estimate the channels by exploiting channel reciprocity in a time-division duplex (TDD) system.
	In this case, the BS sends the pilot signals, and all $K$ devices can simultaneously estimate their channels, including the LoS paths and the RIS-reflected paths. The devices feed back their estimated channels to the BS. 
	The signaling is $\tau L+K \rho N_t N_r M^2$ (symbols), where $\rho$ is the number of symbols to quantize each of the channels and $K{\rho}N_tN_rM^2$ accounts for the feedback of the estimated channel.
	As shown in~\cite{Zhou2021Joint} and~\cite{de2021Channel}, the normalized MSE (NMSE) of the estimated channels is as small as $1.5 \times 10^{-4}$ when $\tau=4$ and $L=100$. When $L$ is sufficiently long, the estimated CSI is close to be perfect.

	Suppose that the transmit symbol ${s_k}$ yields the zero-mean Gaussian distribution with unit variance, and is independent and identically distributed (i.i.d.) between the devices, i.e., $\mathbb{E}[s_ks_{k'}]=0, \forall {k'} \neq k$.
	When perfect CSI is considered, the received superposition signal ${ {\bf{y}}}$ is given by
	\begin{align}
		\label{signal_model_perfect_CSI}
		{ {\bf{y}}} ={}& {\sum\nolimits_{k = 1}^K { \left( { {\bf{H}}_{d,k}} + { {\bf{G}}^{\rm H}}{\bf{\Theta}}{ {\bf{H}}_{r,k}} \right) {{\bf{a}}_k} {s_k}} + {\bf{n}}},
	\end{align}
	where ${{\bf{a}}_k} \in {{\mathbb{C}}^{{N_t} \times 1}}$ is the transmit beamformer of the $k$-th device.
	${\bf{n}} \in {{\mathbb{C}}^{{N_r} \times 1}}$ is the additive white Gaussian noise (AWGN) of the BS, i.e., ${\bf{n}} \sim \mathcal{CN}(0, {{{\sigma}^2_n}}{{\bf{I}}_{N_r}} )$.
	${{{\sigma}^2_n}}$ is the noise power.
	For notational brevity, we define ${ {\bf{H}}_k} \buildrel \Delta \over ={ {\bf{H}}_{d,k}} + { {\bf{G}}^{\rm H}}{\bf{\Theta}}{ {\bf{H}}_{r,k}}$.
	When the channel estimation errors are non-negligible, the received signals can be distorted at the BS.
	It is practical to consider imperfect CSI and develop robust design for the AirFL system.
	The Gaussian-Kronecker model~\cite{Rong2011Robust} is employed to characterize the imperfect estimation of the channels, i.e., ${\bf{H}}_{d,k}={\widehat {\bf{H}}_{d,k}} +  {\Delta {\bf{H}}_{d,k}}$, ${\bf{H}}_{r,k}={\widehat {\bf{H}}_{r,k}} + {\Delta {\bf{H}}_{r,k}}$ and ${\bf{G}}={\widehat {\bf{G}}} + {\Delta {\bf{G}}}$, where ${\widehat {\bf{H}}_{d,k}}$, ${\widehat {\bf{H}}_{r,k}}$ and ${\widehat {\bf{G}}}$ denote the estimated channels, and $\Delta {\bf{H}}_{d,k}$, $\Delta {\bf{H}}_{r,k}$ and ${\Delta {\bf{G}}}$ are the estimation errors with i.i.d. CSCG random entries.
	The estimation errors yield~\cite{Zeng2020joint,Nosrat2011MIMO}
	\begin{align}
		&{\Delta {\bf{H}}_{d,k}} \sim{} \mathcal{CN}\left( {{\bf{0}}_{{N_r} \times {N_t}}}, {{a^2_{d,k}}{{\bf{I}}_{N_t}}} \otimes {{b^2_{d,k}}{{\bf{I}}_{N_r}}} \right), \forall k \in \mathcal{K}, \notag \\
		&{\Delta {\bf{H}}_{r,k}} \sim{} \mathcal{CN}\left( {{\bf{0}}_{{M} \times {N_t}}}, {{a^2_{r,k}}{{\bf{I}}_{N_t}}} \otimes {{b^2_{r,k}}{{\bf{I}}_{M}}} \right), \forall k \in \mathcal{K}, \\
		&{\Delta {\bf{G}}} \sim{} \mathcal{CN}\left( {{\bf{0}}_{{M} \times {N_r}}}, {{a^2_{g}{{\bf{I}}_{N_r}}}} \otimes {{b^2_{g}}{{\bf{I}}_{M}}} \right), \notag
	\end{align}
	where ${a^2_{d,k}}{b^2_{d,k}} = \sigma _{d,k}^2$, ${a^2_{r,k}}{b^2_{r,k}} = \sigma _{r,k}^2$ and ${a^2_{g}}{b^2_{g}} = \sigma _{g}^2$ are the variances of the estimation errors.

	Under imperfect CSI, the received superposition signal ${\widetilde {\bf{y}}}$ at the BS is given by
	\begin{align}
		\label{receive_signal_imperfect}
		{\widetilde {\bf{y}}} =& \sum\nolimits_{k = 1}^K \left[ ( {\widehat {\bf{H}}_{d,k}} + {\Delta {\bf{H}}_{d,k}} ) \notag \right.\\ 
		&\left. + { {( {\widehat {\bf{G}}} + {\Delta {\bf{G}}} )}^{\rm H} {\bf{\Theta}} ( {\widehat {\bf{H}}_{r,k}} + {\Delta {\bf{H}}_{r,k}} ) } \right] {{\bf{a}}_k}{s_k} + {\bf{n}} \notag \\
		=& {\sum\nolimits_{k = 1}^K { {\widehat {\bf{H}}_k} {{\bf{a}}_k} {s_k}}} + \underbrace{{\sum\nolimits_{k = 1}^K { \Delta{\bf{H}}_k}{{\bf{a}}_k}{s_k}}}_{\text{interference~due~to~imperfect~CSI}} + {\bf{n}},
	\end{align}
	where ${\widehat {\bf{H}}_k} \buildrel \Delta \over ={\widehat {\bf{H}}_{d,k}} + {\widehat {\bf{G}}^{\rm H}}{\bf{\Theta}}{\widehat {\bf{H}}_{r,k}}$ and $\Delta {\bf{H}}_k \buildrel \Delta \over = {\Delta {\bf{H}}_{d,k}} + {{\widehat {\bf{G}}}^{\rm H}}{\bf{\Theta}}{\Delta {\bf{H}}_{r,k}} + {{{\Delta {\bf{G}}}^{\rm H}}{{\bf{\Theta}}}{{\widehat {\bf{H}}_{r,k}}}} + {{{\Delta {\bf{G}}}^{\rm H}}{{\bf{\Theta}}}{\Delta {\bf{H}}_{r,k}}}$.

	The integrity of AirFL is susceptible to model poisoning attacks because AirFL directly aggregates the local models and the individual local models are obscure to the BS.
	According to~\cite{Nguyen2021Privacy}, poisoned models have different statistical characteristics from normal models.
	In this sense, it is important to allow the BS to recover the local models and assess their statistics.

	We propose that the BS produces two receive beamformers, ${\bf{b}} \in {\mathbb{C}}^{{N_r} \times 1}$ and ${\bf{f}} \in {\mathbb{C}}^{{N_r} \times 1}$, to aggregate the local models and recover the local models, respectively; see Fig.~\ref{system_model_figure}. 
	%
	The local models are recovered one after another by running SIC.
	The SIC is typically performed in the digital baseband at the BS. Specifically, the analog signals superposed by AirComp are downconverted to the baseband and digitized before the SIC is carried out.
	Considering the effectiveness of the model recovery, we take the convention of SIC that allows the devices with stronger channel gains to be decoded earlier and canceled, so on so forth until all devices are decoded~\cite{ZengM2019Energy},~\cite{Liu2017Energy}. This is because the large-scale path loss typically has a strong impact on the received signal strengths at the BS and, in turn, on the SIC order~\cite{Tabassum2017Modeling}. The BS determines the SIC orders based on the Frobenius norm of the channels of the devices under perfect CSI, i.e., ${\|{\bf{H}}_k\|}^2_F$.
	We assume the devices are ordered (and therefore detected) in the descending order of the Frobenius norms~\cite{Zeng2019Energy,Ni2022Integrating}, i.e., ${\| { {\bf{H}}_1} \|}^2_F \ge {\|  {\bf{H}}_{2} \|}^2_F \ge \cdot \cdot \cdot \ge {\| { {\bf{H}}_K} \|}^2_F$.
	Since the channel estimation errors are agnostic in practice, the SIC orders depend on the Frobenius norms of the estimated channels under imperfect CSI, i.e., ${\|\hat{{\bf{H}}}_k\|}^2_F$.

	Note that the model aggregation and model recovery are in parallel in the proposed framework. The BS can run the model recovery while the devices are training their local models. Alternatively, the BS can choose to recover and examine the local models once a while or only when needed.
	By following the proposed algorithms, each individual model can be recovered at the BS and their trustworthiness can be evaluated using, e.g., the Krum algorithm~\cite{Blanchard2017Machine}, Byzantine-resilient secure aggregation framework~\cite{So2021Byzantine}, or double-masking protocol~\cite{Xu2020VerifyNet}. Misbehaved devices can be identified, held accountable, and suspended from participating in the AirFL.
	
	\section{Beamforming Design and RIS Configuration under Perfect CSI} \label{solution_perfect_CSI}
	In this paper, we minimize the MSE of the aggregated AirFL model while retaining the recoverability of the local models, first under perfect CSI in this section and then under imperfect CSI (as will be described in Section~\ref{solution_imperfect_CSI}). Our design under perfect CSI lays the fundamental design framework with balanced consideration of model accuracy and integrity. With the significant progress made on channel estimation techniques, e.g.,~\cite{Guan2021Anchor,Zhou2021Joint,de2021Channel}, the NMSE between the estimated and actual channel can be reduced as small as $10^{-5}$~\cite{Guan2021Anchor}. In this sense, the consideration of the perfect CSI would not be insubstantial.

	By utilizing the receive beamformer $\bf{b}$ to detect the received signal in (\ref{signal_model_perfect_CSI}), the superposition signal of the aggregated AirFL model is given by
	\vspace{-0.2 cm}
	\begin{equation}
		\label{signal_b_perfect_CSI}
		{\hat s}_b = {\sum\nolimits_{k = 1}^K {{\bf{b}}^{\rm H}{ {\bf{H}}_k}{{\bf{a}}_k} {s_k}}} + {\bf{b}}^{\rm H}{\bf{n}}.
	\end{equation}

	\noindent Under the perfect CSI, the MSE between ${\hat s}_b$ and the desired aggregated model $s$ is given by
	\begin{align}
		\label{MSE_perfect}
		{\rm MSE}\left( {\hat s}_b,s \right) &= \mathbb{E}\!\left[{\left( {\hat s}_b - s \right)}^{\rm H}{\left( {\hat s}_b - s \right)}\right] \notag \\ 
		&= {\sum\nolimits_{k = 1}^K {\left|{{\bf{b}}^{\rm H}}{ {\bf{H}}_k}{{\bf{a}}_k} - 1\right|}^2} + {{\sigma}^2_n}{\left\| {\bf{b}} \right\|^2}.
	\end{align}

	By applying the receive beamformer $\bf{f}$ to the received signal in (\ref{signal_model_perfect_CSI}), the resulting signal for serially recovering the local models is given by
	\begin{equation}
		\label{signal_f_perfect_CSI}
		{\hat s}_f = {\sum\nolimits_{k = 1}^K {{\bf{f}}^{\rm H}{ {\bf{H}}_k}{{\bf{a}}_k} {s_k}}} + {\bf{f}}^{\rm H}{\bf{n}}.
	\end{equation}
	Since the signals recovered prior to the $k$-th device's signal have been subtracted from ${\hat s}_f$, the SINR of the $k$-th device under perfect CSI can be written as
	\begin{equation}
		\label{SINR_perfect_CSI}
		{\hat{\gamma}_k} = \frac{{{{\left| {{{\bf{f}}^{\rm H}}{{ {\bf{H}}}_k}{{\bf{a}}_k}} \right|}^2}}}{{\sum\nolimits_{k' = k + 1}^K {{{\left| {{{\bf{f}}^{\rm H}}{{ {\bf{H}}}_{k'}}{{\bf{a}}_{k'}}} \right|}^2} + \sigma _n^2{{\left\| {\bf{f}} \right\|}^2}} }},\forall k \in \mathcal{K}.
	\end{equation}
	For effective recovery of the local models after the post-processing with $\bf{f}$, the signals recovered successively need to have sufficient power gaps~\cite{MD2016Dynamic}, i.e.,
	\begin{equation}
		\label{perfect_SIC_power_difference}
		{\left| {{{\bf{f}}^{\rm H}}{{ {\bf{H}}_k}}{{\bf{a}}_k}} \right|^2} \!-\! \sum\nolimits_{k' = k + 1}^K \! {{{\left| {{{\bf{f}}^{\rm H}}{{ {\bf{H}}_{k'}}}{{\bf{a}}_{k'}}} \right|}^2}} \!\! \ge \! {{\hat p}_{\rm gap}},\forall k \in \mathcal{K}\backslash \{ K\},\!\!\!
	\end{equation}

	\noindent where ${{\hat p}_{\rm gap}}$ denotes the required minimum power gap between the signal being recovered and those to be recovered.

	Consider that the loss function, e.g., the cross-entropy function~\cite{Wang2022Federated}, decreases with the increase of the correct output probability of each training sample.  
	Reducing the MSE of the aggregated model helps decrease the loss function value~\cite{Cao2022Transmission}.
	In this sense, a smaller MSE is more likely to produce a higher accuracy of AirFL~\cite{Ni2021FL,Xu2021Learning}.
	For this reason, we minimize the MSE of the aggregated AirFL model and retain the recoverability of the local models, by jointly optimizing the transmit beamformers $\{{\bf{a}}_k\}$ at the devices, the receive beamformers $\bf{b}$ and $\bf{f}$ at the BS, and the phase shift matrix $\bf{\Theta}$ of the RIS.
	The problem is cast as
	\vspace{-0.2 cm}
	\begin{subequations}
		\begin{eqnarray}
			\label{p_1_MSE}
			&\mathop {\min }\limits_{{\bf{b}},{\bf{f}},{\bf{\Theta}},\atop{\{{{\bf{a}}_k}\}}} & {\sum\nolimits_{k = 1}^K {\left|{{\bf{b}}^{\rm H}}{ {\bf{H}}_k}{{\bf{a}}_k} - 1\right|}^2} + {{\sigma}^2_n}{\left\| {\bf{b}} \right\|^2} \\
			\label{transmit_power_constraints_perfect_CSI}
			&{\rm s.t}.&{\| {{\bf{a}}_k} \|}^2 \le {P_{\max }}, \forall k \in {\mathcal{K}}, \\
			\label{phase_shift_constraints_perfect_CSI}
			&{}&0 \le {\phi _m} < 2\pi , \forall m \in \mathcal{M}, \\
			\label{QoS_constraints_perfect_CSI}
			&{}&{\hat{\gamma}_k} \ge {\gamma_{\min}}, \forall k \in {\mathcal{K}}, \\
			&{}&\text{(\ref{perfect_SIC_power_difference})}, \notag
		\end{eqnarray}
		\label{p_1}
	\end{subequations}
	
	\vspace{-0.6 cm}
	\noindent where $P_{\max}$ specifies the maximum transmit power of the devices and ${\gamma_{\min}}$ specifies the required minimum SINR of the recovered local models.
	Constraints (\ref{transmit_power_constraints_perfect_CSI}) and (\ref{phase_shift_constraints_perfect_CSI}) specify the ranges for the transmit power of the devices and the phase shifts of the RIS.
	(\ref{QoS_constraints_perfect_CSI}) and (\ref{perfect_SIC_power_difference}) ensure that each local model is recovered with sufficient SINR for effective statistical analysis.

	Problem (\ref{p_1}) has a quadratic objective (\ref{p_1_MSE}) and contraints (\ref{perfect_SIC_power_difference}), (\ref{transmit_power_constraints_perfect_CSI}) and (\ref{QoS_constraints_perfect_CSI}), and is non-convex because of the non-convexity of (\ref{perfect_SIC_power_difference}) and (\ref{QoS_constraints_perfect_CSI}).
	We invoke the AO method to decompose problem (\ref{p_1}) into four subproblems regarding $\bf{b}$, $\{{\bf{a}}_k\}$, $\bf{f}$ and $\bf{\Theta}$.
	A solution with acceptable accuracy and complexity is obtained by solving the subproblems in an alternating manner. 
	
	\subsection{Receive Beamformer for Model Aggregation}
	\label{receive_computation_beamforming_vector_perfect}
	Given fixed transmit beamformers $\{{\bf{a}}_k\}$, receive beamformer $\bf{f}$, and phase shift matrix $\bf{\Theta}$, problem (\ref{p_1}) reduces to a subproblem regarding the receive beamformer $\bf{b}$.
	Since the constraints of problem (\ref{p_1}) are independent of $\bf{b}$, the subproblem is unconstrained, as given by
	\begin{equation}
		\mathop {\min }\limits_{{\bf{b}}} ~ {{\bf{b}}^{\rm{H}}} \! \left( {\sum\limits_{k = 1}^K {{{{\bf{\bar H}}}_{a,k}}} \! + \! \sigma _n^2{{\bf{I}}_{{N_r}}}} \! \right) \! {\bf{b}} \! - \! 2{\mathop{\rm Re}\nolimits} \left\{ \! {{{\bf{b}}^{\rm{H}}}\sum\limits_{k = 1}^K {{{ {\bf{H}}}_k}{{\bf{a}}_k}} } \! \right\}
		\label{p_2}
	\end{equation}
	where ${{{{\bf{\bar H}}}_{a,k}}} = { {\bf{H}}_k}{{\bf{a}}_k}{{\bf{a}}_k^{\rm H}}{ {\bf{H}}_k^{\rm H}}$.
	Problem (\ref{p_2}) is convex. 
	We can obtain the closed-form solution by following the MMSE rule to evaluate the first-order derivative with respect to (w.r.t.) $\bf{b}$:
	\begin{equation}
		\label{optimal_b_perfect_CSI}
		{\bf{b}} = {\left( {\sum\nolimits_{k = 1}^K {{{{\bf{\bar H}}}_{a,k}}}  + \sigma _n^2{{\bf{I}}_{{N_r}}}} \right)}^{-1}\left(\sum\nolimits_{k = 1}^K {{{ {\bf{H}}}_k}{{\bf{a}}_k}} \right).
	\end{equation}
	
	\subsection{Transmit Beamformer} \label{transmit_beamforming_vectors_perfect}
	Given fixed receive beamformers $\bf{b}$ and $\bf{f}$, and phase shifts $\bf{\Theta}$, the subproblem of $\{{\bf{a}}_k\}$ is
	\begin{subequations}
		\begin{eqnarray}
			&\mathop {\min }\limits_{\{{{\bf{a}}_k}\}} & \sum\nolimits_{k = 1}^K \left\{{{\bf{a}}_k^{\rm{H}}{{\bar {\bf{H}} }_{b,k}}{{\bf{a}}_k}}  - 2{\rm{Re}}\left\{ {{\bf{a}}_k^{\rm{H}} {\bf{H}}_k^{\rm{H}}{\bf{b}}} \right\} \right\} \\
			\label{ak_constraints_1_perfect}
			&{\rm s.t.}& {{\bf{a}}^{\rm H}_k}{{\bf{I}}_{N_t}}{{\bf{a}}_k}  - {P_{\max }} \le 0, \forall k \in {\mathcal{K}}, \\
			\label{p_3_c_2}
			&{}&{-{{{\bf{a}}}_k^{\rm H}}{{\bar {\bf{H}}}_{f,k}}{{\bf{a}}_k}} + {\gamma_{\min}}{\sum\nolimits_{k' = k + 1}^K {{{{\bf{a}}}_{k'}^{\rm H}}{{\bar {\bf{H}}}_{f,{k'}}}{{\bf{a}}_{k'}}}}\notag \\
			&{}&+{{\gamma_{\min}}{\sigma_n^2}\|{\bf{f}}\|^2} \le 0, 	\forall k \in {\mathcal{K}}, \\
			\label{p_3_c_3}
			&{}&{-{{{\bf{a}}}_k^{\rm H}}{{\bar 	{\bf{H}}}_{f,k}}{{\bf{a}}_k}} + {\sum\nolimits_{k' = k + 1}^K {{{{\bf{a}}}_{k'}^{\rm H}}{{\bar {\bf{H}}}_{f,{k'}}}{{\bf{a}}_{k'}}}}\notag \\
			&{}&+{{\hat p}_{\rm gap}} \le 0, \forall k \in \mathcal{K}\backslash \{ K\},
		\end{eqnarray}
		\label{p_3}
	\end{subequations}

	\vspace{-0.5 cm}
	\noindent where ${{{{\bf{\bar H}}}_{b,k}}} = { {\bf{H}}^{\rm H}_k}{\bf{b}}{{\bf{b}}^{\rm H}}{ {\bf{H}}_k}$ and ${{{{\bf{\bar H}}}_{f,k}}} = { {\bf{H}}^{\rm H}_k}{\bf{f}}{{\bf{f}}^{\rm H}}{ {\bf{H}}_k}$.
	Problem (\ref{p_3}) is a non-convex quadratically constrained quadratic program (QCQP) due to the concave terms in (\ref{p_3_c_2}) and (\ref{p_3_c_3}).
	
	%
	We first expand ${\bf{a}}_k$ to ${\bar {\bf{a}}_k} = [{\bf{a}}^{\rm H}_k,u^{\rm H}_k]^{\rm H}$ with auxiliary variables $\{u_k\}_{\forall k \in \mathcal{K}}$, $u_k \in \mathbb{R}$ and $u^2_k=1$, and then define a matrix ${\bf{A}}_k \buildrel \Delta \over ={\bar {\bf{a}}_k}{\bar {\bf{a}}^{\rm H}_k},\forall k \in \mathcal{K}$. 
	%
	To write problem (\ref{p_3}) in a matrix form, we define
	\begin{align}
		&{{\bf{Z}}_{0,k}}\!\!\!\! & \! \buildrel \Delta \over = {\hspace{-0.2 cm}} \left[ {\begin{array}{*{20}{c}} {{{{\bf{\bar H}}}_{b,k}}}&{ -  {\bf{H}}_k^{\rm{H}}{\bf{b}}}\\
				{ - {{\bf{b}}^{\rm{H}}}{{ {\bf{H}}}_k}}&0 \end{array}} \right]\!,
		{{\bf{Z}}_{1,k}} & \!  \buildrel \Delta \over = {\hspace{-0.2 cm}} \left[ {\begin{array}{*{20}{c}} {{{\bf{I}}_{{N_t}}}}&{{{\bf{0}}_{{N_t} \times 1}}}\\
				{{\bf{0}}_{{N_t} \times 1}^{\rm{H}}}&0 \end{array}} \right]\!, \notag \\
		&{{\bf{Z}}_{2,k}}\!\!\!\! & \! \buildrel \Delta \over = {\hspace{-0.2 cm}} \left[ {\begin{array}{*{20}{c}} {{{{\bf{\bar H}}}_{f,k}}}&{{{\bf{0}}_{{N_t} \times 1}}}\\
				{{\bf{0}}_{{N_t} \times 1}^{\rm{H}}}&0 \end{array}} \right]\!, \forall k \! \in \! \mathcal{K}.
	\end{align}
	Problem (\ref{p_3}) is recast as a semidefinite program (SDP), as given by
	\vspace{-0.2 cm}
	\begin{subequations}
		\begin{eqnarray}
			\label{SDR_objective_perfect}
			&\mathop {\min }\limits_{\{{{\bf{A}}_k}\}} & \sum\nolimits_{k = 1}^K {\text{tr}({{\bf{Z}}_{0,k}}{{\bf{A}}_k})} \\
			\label{SDR_constraints_1_perfect}
			&{\rm s.t}.& \text{tr}({{\bf{Z}}_{1,k}}{{\bf{A}}_k})  - {P_{\max }} \le 0, \forall k \in {\mathcal{K}}, \\
			\label{SDR_constraints_2_perfect}
			&{}&-\text{tr}({{\bf{Z}}_{2,k}}{{\bf{A}}_k}) + {\gamma_{\min}}{\sum\nolimits_{k' = k + 1}^K {\text{tr}({{\bf{Z}}_{2,k'}}{{\bf{A}}_{k'}})}} \notag \\
			&{}&+{{\gamma_{\min}}{\sigma_n^2}\|{\bf{f}}\|^2} \le 0, \forall k \in {\mathcal{K}}, \\
			\label{SDR_constraints_3_perfect}
			&{}&-\text{tr}({{\bf{Z}}_{2,k}}{{\bf{A}}_k}) + {\sum\nolimits_{k' = k + 1}^K {\text{tr}({{\bf{Z}}_{2,k'}}{{\bf{A}}_{k'}})}}\notag \\
			&{}&+{{\hat p}_{\rm gap}} \le 0, \forall k \in \mathcal{K}\backslash \{ K\}, \\
			\label{SDR_constraints_4_perfect}
			&{}&[{\bf{A}}_k]_{N_t+1,N_t+1}=1, \forall k \in \mathcal{K}, \\
			\label{SDR_constraints_5_perfect}
			&{}&{\bf{A}}_k \succeq 0, \forall k \in \mathcal{K}, \\
			\label{SDR_constraints_6_perfect}
			&{}&\text{rank}({\bf{A}}_k) = 1,\forall k \in \mathcal{K}.
		\end{eqnarray}
		\label{p_4}
	\end{subequations}
	
	\vspace{-0.6 cm}
	\noindent Problem (\ref{p_4}) is non-convex because of the rank constraint~(\ref{SDR_constraints_6_perfect}).
	%
	%
	We invoke DC programming~\cite{Yang2020Federated} to solve (\ref{p_4}), where (\ref{SDR_constraints_6_perfect}) is equivalently rewritten as
	\begin{equation}
		\label{equavalent_rank_constraints}
		\text{tr}({\bf{A}}_k)-\|{\bf{A}}_k\|_2=0, \forall k \in \mathcal{K}.
	\end{equation}

	\noindent The equivalence between (\ref{SDR_constraints_6_perfect}) and (\ref{equavalent_rank_constraints}) is due to the fact that $\text{tr}({\bf{A}}_k) = \|{\bf{A}}_k\|_2 = \omega_{{\bf{A}}_k}$ if $\text{rank}({\bf{A}}_k) = 1$, where $\omega_{{\bf{A}}_k}$ is the maximum singular value of ${\bf{A}}_k$.
	By replacing (\ref{SDR_constraints_6_perfect}) with (\ref{equavalent_rank_constraints}) and making it as the regularizer in (\ref{SDR_objective_perfect}), problem (\ref{p_4}) becomes a DC programming:
	\begin{eqnarray}
		\label{p_5}
		\hspace{-0.6 cm}&\mathop {\min }\limits_{\{{{\bf{A}}_k}\}} &\sum\nolimits_{k = 1}^K \left\{\text{tr}({{\bf{Z}}_{0,k}}{{\bf{A}}_k}) + \alpha \left(\text{tr}({{\bf{A}}_k})-\|{\bf{A}}_k\|_2\right)\right\}\\
		\hspace{-0.6 cm}&{\rm s.t.}& \text{(\ref{SDR_constraints_1_perfect})}-\text{(\ref{SDR_constraints_5_perfect})}, \notag
	\end{eqnarray}
	
	\vspace{-0.2 cm}
	\noindent where $\alpha$ is a penalty factor.
	Problem (\ref{p_5}) is still non-convex due to the 2-norm $\|{\bf{A}}_k\|_2$.

	We linearize $\|{\bf{A}}_k\|_2$ by using its linearization $\langle {{\bf{\dot A}}}^{(t)}_k , {\bf{A}}_k \rangle = \text{tr}(({{\bf{\dot A}}}^{(t)}_k)^{\rm H}{{\bf{A}}_k})$, where ${{\bf{A}}}^{(t)}_k$ is obtained at the $t$-th iteration of the	 DC programming, ${{\bf{\dot A}}}^{(t)}_k={\bf{u}}^{(t)}_k({\bf{u}}^{(t)}_k)^{\rm H}$ is a subgradient of $\|{\bf{A}}_k\|_2$ at ${{\bf{A}}}^{(t)}_k$, and ${\bf{u}}^{(t)}_k$ is the singular vector associated with $\omega_{{{\bf{A}}}^{(t)}_k}$~\cite{G1992Characterization}.
	As a result, problem (\ref{p_5}) is convexified w.r.t. $\{{\bf{A}}_k\}$, as given by
	\begin{eqnarray}
		\label{p_6}
		\hspace{-0.6 cm}&\mathop {\min }\limits_{\{{{\bf{A}}_k}\}} &\sum\nolimits_{k = 1}^K \left\{\text{tr}(({{\bf{Z}}_{0,k}}+\alpha {\bf{I}}_{N_t}){{\bf{A}}_k}) - \alpha \langle {{\bf{\dot A}}}^{(t)}_k , {\bf{A}}_k \rangle\right\} \\
		\hspace{-0.6 cm}&{\rm s.t.}&\text{(\ref{SDR_constraints_1_perfect})}-\text{(\ref{SDR_constraints_5_perfect})}, \notag
	\end{eqnarray}
	
	\vspace{-0.2 cm}
	\noindent which can be solved by CVX toolkits~\cite{Grant2014CVX}.
	A rank-one solution for $\{{\bf{A}}_k\}$ is obtained by solving (\ref{p_6}) iteratively.
	$\bar{\bf{a}}_k$ is obtained by eigenvalue decomposition, i.e., $\bar{\bf{a}}_k = \sqrt {{\lambda _{{\bf{A}}_k}}} {{\bf{p}}_{k}},\forall k \in \mathcal{K}$.
	${\lambda _{{\bf{A}}_k}}$ is the largest eigenvalue of ${\bf{A}}_k$.
	${{\bf{p}}_{k}}$ is the corresponding eigenvector.
	The solution of ${\bf{a}}_k$ is obtained by removing the last element of $\bar{\bf{a}}_k$, $\bar{\bf{a}}_k=[{\bf{a}}^{\rm H}_k,u^{\rm H}_k]^{\rm H}$ with $u^2_k=1$.

	\vspace{-0.2 cm}
	\subsection{Receive Beamformer for Local Model Recovery} \label{receive_decoding_beamforming_perfect}
	The objective (\ref{p_1_MSE}) is independent of the receive beamformer $\bf{f}$.
	Given fixed receive beamformer $\bf{b}$, transmit beamformers $\{{\bf{a}}_k\}$, and phase shift matrix $\bf{\Theta}$, finding $\bf{f}$ is a feasibility problem:
	\vspace{-0.2 cm}
	\begin{subequations}
		\begin{eqnarray}
			&\mathop {\text{find} }\limits_{{\bf{f}}} & {\bf{f}} \\
			\label{f_constraints_1}
			&{\rm s.t.}& {\bf{f}}^{\rm H}{{\bf{B}}_{1,k}}{\bf{f}} \le 0 , \forall k \in {\mathcal{K}}, \\
			\label{f_constraints_2}
			&{}&{\bf{f}}^{\rm H}{{\bf{B}}_{2,k}}{\bf{f}} + {{{\hat p}_{gap}}} \le 0 , \forall k \in {\mathcal{K}}\backslash \{ K\},
		\end{eqnarray}
		\label{p_7}
	\end{subequations}
	
	\vspace{-0.6 cm}
	\noindent where ${{\bf{B}}_{1,k}}$ and ${{\bf{B}}_{2,k}}$ are defined as
	\vspace{-0.2 cm}
	\begin{align}
		&{{\bf{B}}_{1,k}} \! \buildrel \Delta \over  =\! {\gamma _{\min }} (\sum\nolimits_{k' = k + 1}^K \! {{{\bar {\bf{H}}}_{a,k'}}}\!+\!\sigma _n^2{{\bf{I}}_{{N_r}}}) \!-\! {{\bar {\bf{H}}}_{a,k}}, \forall k \in \mathcal{K}, \\
		&{{\bf{B}}_{2,k}} \! \buildrel \Delta \over =\! \sum\nolimits_{k' = k + 1}^K {{{\bar {\bf{H}}}_{a,k'}}} \!-\! {{\bar {\bf{H}}}_{a,k}}, \forall k \in \mathcal{K}\backslash \{ K\}.
	\end{align}

	In light of~\cite{Liu2021Intelligent}, we transform problem (\ref{p_7}) to a minimization problem w.r.t. an auxiliary variable $\beta \in \mathbb{R}$, as given by
	\vspace{-0.6 cm}
	\begin{subequations}
		\begin{eqnarray}
			&\mathop {\min }\limits_{{\bf{f}},{\beta \le 0}} & {\beta} \\
			\label{constrain_1_f}
			&{\rm s.t.}& {\bf{f}}^{\rm H}{{\bf{B}}_{1,k}}{\bf{f}} \le {\beta} , \forall k \in {\mathcal{K}}, \\
			\label{constrain_2_f}
			&{}&{\bf{f}}^{\rm H}{{\bf{B}}_{2,k}}{\bf{f}} + {{{\hat p}_{gap}}} \le {\beta} , \forall k \in {\mathcal{K}}\backslash \{ K\}.
		\end{eqnarray}
		\label{p_8}
	\end{subequations}
	
	\vspace{-0.6 cm}
	\noindent Problem (\ref{p_8}) reinforces the local model recoverability by requiring higher SINRs and larger power gaps than the original problem (\ref{p_7}).
	Since ${{\bf{B}}_{1,k}}$ and ${{\bf{B}}_{2,k}}$ are indefinite, (\ref{p_8}) is non-convex.
	We employ the SCA to convexify (\ref{p_8}), where a sequence of feasible points are generated by minimizing convex surrogate functions until convergence.
	The two quadratic surrogate functions of (\ref{constrain_1_f}) and (\ref{constrain_2_f}) are given by~\cite{Sun2017Majorization,Zheng2022Semi}:
	\vspace{-0.3 cm}
	\begin{align}
		\label{surrogate_Ak}
		&{\hat g_{1,k}}\left( {\bf{f}} | {{\bf{f}}^{(t)}} \right) = {} {\left({{\bf{f}}^{(t)}}\right)^{\rm{H}}}{{\bf{B}}_{1,k}}{{\bf{f}}^{(t)}}\!+\! {\omega _{{{\bf{B}}_{1,k}}}}\left\|{{\bf{f}}  -  {{\bf{f}}^{(t)}}}\right\|^2  \notag \\ 
		&+ 2{\mathop{\rm Re}\nolimits} \left\{ {{{\left({{\bf{B}}_{1,k}}{{\bf{f}}^{(t)}}\right)}^{\rm{H}}}\left({\bf{f}}  -  {{\bf{f}}^{(t)}}\right)} \right\}, \forall k \! \in \! {\mathcal{K}}, \\
		\label{surrogate_Bk}
		&{\hat g_{2,k}}\left( {\bf{f}} | {{\bf{f}}^{(t)}} \right) = {} {\left( {{{\bf{f}}^{(t)}}} \right)^{\rm{H}}}{{\bf{B}}_{2,k}}{{\bf{f}}^{(t)}} \!+\!  {\omega _{{{\bf{B}}_{2,k}}}} \left\|{ {{\bf{f}}  -  {{\bf{f}}^{(t)}}} }\right\|^2 \notag \\
		&+ 2{\mathop{\rm Re}\nolimits} \left\{ {{{\left( {{{\bf{B}}_{2,k}}{{\bf{f}}^{(t)}}} \right)}^{\rm{H}}}\left( {{\bf{f}}  -  {{\bf{f}}^{(t)}}} \right)} \right\}, \forall k \in {\mathcal{K}}\backslash \{K\},
	\end{align}
	
	\vspace{-0.2 cm}
	\noindent where $\omega _{{{\bf{B}}_{1,k}}}$ and $\omega _{{{\bf{B}}_{2,k}}}$ are the maximum singular values of ${\bf{B}}_{1,k}$ and ${\bf{B}}_{2,k}$, respectively; and ${\bf{f}}^{(t)}$ is the result of $\bf{f}$ obtained at the $t$-th iteration of the SCA.
	As a result, solving (\ref{p_8}) becomes iteratively solving the problem below.
	\vspace{-0.2 cm}
	\begin{subequations}
		\begin{eqnarray}
			&\mathop {\min }\limits_{{\bf{f}},{\beta \le 0}} & {\beta} \\
			&{\rm s.t.}& {\hat g_{1,k}}\left( {\bf{f}} | {{\bf{f}}^{(t)}} \right) \le {\beta} , \forall k \in {\mathcal{K}}, \\
			&{}&{\hat g_{2,k}}\left( {\bf{f}} | {{\bf{f}}^{(t)}} \right)  + {{{\hat p}_{gap}}} \le {\beta} , \forall k \in {\mathcal{K}}\backslash \{ K\},
		\end{eqnarray}
		\label{p_9}
	\end{subequations}
	
	\vspace{-0.5 cm}
	\noindent which is convex in $\{{\bf{f}},{\beta}\}$ and solved using CVX toolkits.
	
	\subsection{Phase Shift Matrix} \label{phase_shift_matrix_perfect_CSI}
	Given fixed transmit and receive beamformers, i.e., $\{{\bf{a}}_k\}$, $\bf{b}$ and $\bf{f}$, we reformulate problem (\ref{p_1}) concerning $\bf{\Theta}$ to a problem concerning the phase shifts ${\bf{v}} \buildrel \Delta \over = {[{{\phi}_1},...,{{\phi}_M}]}^{\rm T}$.
	We employ the SCA method to solve $\bf{v}$~\cite{Zeng2020joint}.
	The problem regarding $\bf{v}$ is reconstructed as follows.

	Given fixed receive beamformers $\bf{b}$ and $\bf{f}$, and transmit beamformers $\{{\bf{a}}_k\}$, problem (\ref{p_1}) can be equivalently rewritten as the following problem w.r.t. the phase shifts $\bf{v}$:
	\vspace{-0.2 cm}
	\begin{subequations}
		\begin{eqnarray}
			&\mathop {\min }\limits_{ {\bf{v}} } & {{h_0}\left( {\bf{v}} \right)} \\
			&{\rm s.t.}& {h_{1,k}}\left( {\bf{v}} \right) \le 0 , \forall k \in {\mathcal{K}}, \\
			&{}&{h_{2,k}}\left( {\bf{v}} \right) \le 0 , \forall k \in {\mathcal{K}}\backslash \{ K\},
		\end{eqnarray}
		\label{p_10}
	\end{subequations}
	\!\!where the notations are defined in Table~\ref{table_I}. 
	The derivation for problem transformation is given in Appendix \ref{proof_of_proposition_1}.
	
	The solution to problem (\ref{p_10}) should satisfy (\ref{phase_shift_constraints_perfect_CSI}). Nevertheless, we can drop (\ref{phase_shift_constraints_perfect_CSI}) since $e^{j\phi}$ is a periodic function with the period of $2\pi$. 
	The solution under (\ref{phase_shift_constraints_perfect_CSI}) is the remainder of the Euclidean division of the solution to (\ref{p_10}) by $2\pi$.
	Since the subproblem concerning the phase shifts is a non-convex quadratic constrained quadratic programming (QCQP) problem, we employ the SCA to solve the problem.
	Problem (\ref{p_10}) could also be solved approximately using SDR with the worst-case complexity growing quartically with the number of reflecting elements~\cite{Zhi2010Semidefinite}.
	In contrast, the SCA method solves (\ref{p_10}) with a cubic complexity.
	
	\begin{table}[t]
		\caption{Definitions of Notations used in problem (\ref{p_10}), where ${\rm vec}(\cdot)$ vectorizes the diagonal of a matrix.}
		\label{table_I}
		\centering
			\begin{tabular}{|l|l|}
				\hline
				\textbf{Notation} & \textbf{Definition} \\ \cline{1-2}
				
				\rule{0pt}{12pt}${e^{j{\bf{v}}}}$ & ${\rm vec}({\bf{\Theta}})$  \\[3pt] \cline{1-2}
				
				\rule{0pt}{12pt}${{{\bf{\bar G}}}_b}$, ${{{\bf{\bar G}}}_f}$ & $ {\bf{G}}{\bf{b}}{{\bf{b}}^{\rm{H}}}{ {\bf{G}}^{\rm{H}}}$, $ {\bf{G}}{\bf{f}}{{\bf{f}}^{\rm{H}}}{ {\bf{G}}^{\rm{H}}}$  \\[3pt] \cline{1-2}
				
				\rule{0pt}{12pt}${{\bf{Q}}_{0,k}}$ & ${ {\bf{H}}_{r,k}}{{\bf{a}}_k}{\bf{a}}_k^{\rm{H}} {\bf{H}}_{r,k}^{\rm{H}}$ \\[3pt] \cline{1-2}
				
				\rule{0pt}{12pt}${{\bf{Q}}_{1,k}}$ & \tabincell{l}{ $ {{\bf{b}}^{\rm{H}}}{{ {\bf{H}}}_{d,k}}{{\bf{a}}_k}{\bf{G}}{\bf{ba}}_k^{\rm{H}} {\bf{H}}_{r,k}^{\rm{H}}- {\bf{G}}{\bf{ba}}_k^{\rm{H}} {\bf{H}}_{r,k}^{\rm{H}}$} \\[3pt] \cline{1-2} 
				
				\rule{0pt}{12pt}${{\bf{Q}}_{2,k}}$ & $ {\bf{G}}{\bf{f}}{{\bf{f}}^{\rm{H}}}{ {\bf{H}}_{d,k}}{{\bf{a}}_k}{\bf{a}}_k^{\rm{H}} {\bf{H}}_{r,k}^{\rm{H}}$  \\[3pt] \cline{1-2}
				
				\rule{0pt}{12pt}${{\bf{F}}_0}$ & $ \sum\nolimits_{k = 1}^K {{{{\bf{\bar G}}}_b} \circ {{\bf{Q}}_{0,k}^{\rm T}}}$  \\[3pt] \cline{1-2}
				
				\rule{0pt}{12pt}${{\bf{F}}_{1,k}}$ &  \tabincell{l}{$ - {\gamma _{\min }}\sum\nolimits_{k' = k + 1}^K {{{{\bf{\bar G}}}_f} \circ {{\bf{Q}}^{\rm T}_{0,k'}}}$ $+ {{{\bf{\bar G}}}_f} \circ {{\bf{Q}}^{\rm T}_{0,k}}$}  \\[3pt] \cline{1-2}
				
				\rule{0pt}{12pt}${{\bf{F}}_{2,k}}$ & \tabincell{l}{ $- \sum\nolimits_{k' = k + 1}^K {{{{\bf{\bar G}}}_f} \circ {{\bf{Q}}^{\rm T}_{0,k'}}}$ $+{{{\bf{\bar G}}}_f} \circ {{\bf{Q}}^{\rm T}_{0,k}}$} \\[3pt] \cline{1-2}
				
				\rule{0pt}{12pt}${{\bf{r}}_0}$ & ${\rm vec}(\sum\nolimits_{k=1}^K{{{\bf{Q}}_{1,k}}})$ \\[3pt] \cline{1-2}

				\rule{0pt}{12pt}${{\bf{r}}_{1,k}}$ &\tabincell{l}{ ${{\rm vec}({{{\bf{Q}}_{2,k}}})} - {\gamma _{\min }}\sum\nolimits_{k' = k + 1}^K \! {{{\rm vec}({{{\bf{Q}}_{2,k'}}})}}$} \\[3pt] \cline{1-2}
				
				\rule{0pt}{12pt}${{\bf{r}}_{2,k}}$ & \tabincell{l}{ ${{\rm vec}({{{\bf{Q}}_{2,k}}})} - \sum\nolimits_{k' = k + 1}^K \! {{{\rm vec}({{{\bf{Q}}_{2,k'}}})}}$  } \\[3pt] \cline{1-2}
				
				\rule{0pt}{12pt}${C_{1,k}}$ & \tabincell{l}{${\gamma _{\min }}(\sum\nolimits_{k' = k + 1}^K {|{{\bf{f}}^{\rm{H}}}{{ {\bf{H}}}_{d,k'}}{{\bf{a}}_{k'}}|^2} \! + \! \sigma _n^2{\left\| {\bf{f}} \right\|^2})$ \\ $ - |{{\bf{f}}^{\rm{H}}}{ {\bf{H}}_{d,k}}{{\bf{a}}_k}|^2$}  \\[3pt] \cline{1-2}
				
				\rule{0pt}{12pt}${C_{2,k}}$ & \tabincell{l}{$\sum\nolimits_{k' = k + 1}^K {|{{\bf{f}}^{\rm{H}}}{{ {\bf{H}}}_{d,k'}}{{\bf{a}}_{k'}}|^2} \!+\! {{\hat p}_{\rm gap}} $
					$ \!- |{{\bf{f}}^{\rm{H}}}{ {\bf{H}}_{d,k}}{{\bf{a}}_k}|^2$} \\[3pt] \cline{1-2}
				
				\rule{0pt}{12pt}${h_0}\left( {\bf{v}} \right)$ & ${\left( {{e^{j{\bf{v}}}}} \right)^{\rm{H}}}{{\bf{F}}_0}{e^{j{\bf{v}}}} + 2{\mathop{\rm Re}\nolimits} \{ {{{\left( {{e^{j{\bf{v}}}}} \right)}^{\rm{H}}}{{\bf{r}}_0}} \}$ \\[3pt] \cline{1-2}
				
				\rule{0pt}{12pt}${h_{1,k}}\left( {\bf{v}} \right)$ & \tabincell{l}{$- {\left( {{e^{j{\bf{v}}}}} \right)^{\rm{H}}}{{\bf{F}}_{1,k}}{e^{j{\bf{v}}}} - 2{\mathop{\rm Re}\nolimits} \{ {{{\left( {{e^{j{\bf{v}}}}} \right)}^{\rm{H}}}{{\bf{r}}_{1,k}}} \}$ $+ C_{1,k}$} \\[3pt] \cline{1-2}
				
				\rule{0pt}{12pt}${h_{2,k}}\left( {\bf{v}} \right)$ & \tabincell{l}{$- {\left( {{e^{j{\bf{v}}}}} \right)^{\rm{H}}}{{\bf{F}}_{2,k}}{e^{j{\bf{v}}}} - 2{\mathop{\rm Re}\nolimits} \{ {{{\left( {{e^{j{\bf{v}}}}} \right)}^{\rm{H}}}{{\bf{r}}_{2,k}}} \}$ $+ C_{2,k}$} \\[3pt] \cline{1-2}
				
				\hline
		\end{tabular}
	\end{table}

	The key step of the SCA is to apply the second-order Taylor expansion to approximate the surrogate functions for the QCQP problem, as given by~\cite{Sun2017Majorization}
	\begin{align}
		\label{surrogate_hn}
		{\hat h_l}\left( {\bf{v}} | {{\bf{v}}^{(t)}} \right) ={}& {h_l}\left( {\bf{v}}^{(t)} \right)+ \nabla {h_l}{\left( {{{\bf{v}}^{(t)}}} \right)^{\rm{T}}}\left( {{\bf{v}} - {{\bf{v}}^{(t)}}} \right) \notag \\
		&+ \frac{{{\xi _l}}}{2}{\left\| {{\bf{v}} - {{\bf{v}}^{(t)}}} \right\|^2}, \forall l \in \mathcal{L},
	\end{align}
	where $\mathcal{L} =\{0\}\cup{\{(1,k)\}}\cup{\{(2,k)\}}$, ${{\bf{v}}^{(t)}}$ is the result of $\bf{v}$ from the $t$-th iteration of SCA, $\nabla {h_l}{\left( {{{\bf{v}}}} \right)}$ is the gradient, ${\nabla ^2}{h_l}\left( {\bf{v}} \right)$ is the Hessian, and ${\xi _l}$ is a constant.

	Then, (\ref{p_10}) is convexified and readily solved using CVX toolbox, as given by
	\vspace{-0.2 cm}
	\begin{subequations}
		\begin{eqnarray}
			&\mathop {\min }\limits_{ {\bf{v}} } & {{\hat h_0}\left( {\bf{v}} | {{\bf{v}}^{(t)}} \right)} \\
			&{\rm s.t.}& {\hat h_{1,k}}\left( {\bf{v}} | {{\bf{v}}^{(t)}} \right) \le 0 , \forall k \in {\mathcal{K}}, \\
			&{}&{\hat h_{2,k}}\left( {\bf{v}} | {{\bf{v}}^{(t)}} \right) \le 0 , \forall k \in {\mathcal{K}}\backslash \{ K\}.
		\end{eqnarray}
		\label{p_11}
	\end{subequations}
	\vspace{-0.6 cm}

	It is critical to determine the second-order coefficient ${\xi _l},\forall l \in \mathcal{L}$, to ensure that the surrogate function is an upper bound of the original function in the SCA or, in other words, to ensure ${\xi _l}{{\bf{I}}_M} \succeq {\nabla ^2}{h_l}\left( {\bf{v}} \right), \forall l \in \mathcal{L}$.
	The Armijo rule is often used to determine $\{{\xi _l}\}$~\cite{Zeng2020joint,Guo2020Weighted}.
	However, the Armijo rule has a quadratic complexity here for iterative search of $\{{\xi _l}\}$~\cite{Guo2020Weighted,Bertsekas1999Nonlinear}.
	In contrast, we determine ${\xi _l}$ analytically with a substantially lower complexity, as below.
	\begin{lemma}
		\label{lemma_1}
		We have ${\xi _l}{{\bf{I}}_M} \succeq {\nabla ^2}{h_l}\left( {\bf{v}} \right), \forall l \in \mathcal{L}$ for the Hermitian ${\bf{F}}_l, \forall l \in \mathcal{L}$ defined in Table~\ref{table_I} and constants ${\xi _l}, \forall l \in \mathcal{L}$ satisfying
		\begin{align}
			\label{lemma_2_inequality}
			{\xi _l} \ge{}& 2\mathop {\max }\limits_{i \in {\cal M}} \left\{ {\sum\limits_{j = 1}^M {\left| {{{\left[ {{{\bf{F}}_l}} \right]}_{j,i}}} \right|}  + \left| {{{\left[ {{{\bf{r}}_l}} \right]}_i}} \right|} \right\} + 2{\omega _{{{\bar{\bf{F}}}_l}}} \notag \\
			&+ 2\mathop {\max }\limits_{i \in {\cal M}} \left\{ {\left| {{{\left[ {{{\bf{F}}_l}} \right]}_{i,i}}} \right|} \right\},\forall l \in {\cal L},
		\end{align}
		where ${{\bar{\bf{F}}}_l} = {\bf{F}}_l^{\rm{T}} - {\rm diag}( {{{\left[ {{{\bf{F}}_l}} \right]}_{1,1}},...,{{\left[ {{{\bf{F}}_l}} \right]}_{M,M}}} ), \forall l \in {\cal L}$; ${{{\left[ {{{\bf{F}}_l}} \right]}_{j,i}}}$ is the $(i,j)$-th entry of ${{\bf{F}}_l}$; ${{{\left[ {{{\bf{r}}_l}} \right]}_i}}$ is the $i$-th entry of vector ${{{\bf{r}}_l}}$; and ${\omega _{{{\bar{\bf{F}}}_l}}}$ is the maximum singular value of ${{\bar{\bf{F}}}_l}$.
	\end{lemma}
	\begin{IEEEproof}
		Please refer to Appendix B.
	\end{IEEEproof}

	By following  Lemma \ref{lemma_1}, the constants $\{{\xi _l}\}$ are first determined with ${\bf{F}}_l, \forall l \in \mathcal{L}$ according to (\ref{lemma_2_inequality}), before the SCA starts.
	Then, problem (\ref{p_11}) is constructed and solved.
	Recall that the elements of ${\bf{v}}$ are in $[0,2\pi)$.
	The phase shift matrix ${\bf{\Theta}}$ is obtained by diagonalizing ${e^{j{\bf{v}}}}$.
	%
	
	\begin{algorithm}[t]
			\caption{Proposed Algorithm for Perfect CSI Case}
			\label{algorithm_1}
			\begin{algorithmic}[1]
				\renewcommand{\algorithmicrequire}{\textbf{Initialize}}
				\renewcommand{\algorithmicensure}{\textbf{Output}}
				\STATE \textbf{Initialize} a feasible solution $({{\bf{b}}^{(0)}},\{ {\bf{a}}_k^{(0)} \},{\bf{f}}^{(0)},{\bf{\Theta}}^{(0)})$, the maximum iteration numbers $T_0$, $T_1$, $T_2$, $T_3$, the convergence accuracies $\varepsilon_0$, $\varepsilon_1$, $\varepsilon_2$, $\varepsilon_3$, and set $t=0$, $t'=0$.
				\REPEAT
				\STATE Update $t \leftarrow t+1$.
				\STATE Given $\{ {\bf{a}}_k^{(t-1)} \}$, ${\bf{f}}^{(t-1)}$, ${\bf{\Theta}}^{(t-1)}$, calculate ${{\bf{b}}^{(t)}}$ via (\ref{optimal_b_perfect_CSI}).
				\REPEAT
				\STATE Update $t' \leftarrow t'+1$.
				\STATE Calculate ${{\bf{\dot A}}}^{(t'-1)}_k={\bf{u}}^{(t'-1)}_k({\bf{u}}^{(t'-1)}_k)^{\rm H},\forall k \in \mathcal{K}$.
				\STATE Given ${{\bf{b}}^{(t)}}$, ${\bf{f}}^{(t-1)}$, ${\bf{\Theta}}^{(t-1)}$, obtain $\{{\bf{A}}^{(t')}_k\}$ by solving (\ref{p_6}) using CVX.
				\UNTIL $t' \ge T_1$ or $\text{tr}({\bf{A}}^{(t')}_k)-\|{\bf{A}}^{(t')}_k\|_2 \le \varepsilon_1, \forall k \in \mathcal{K}$;
				\STATE Recover $\bar{\bf{a}}^{(t)}_k = \sqrt {{\lambda _{{\bf{A}}^{(t')}_k}}} {{\bf{p}}_{k}}, \forall k \in \mathcal{K}$.
					\STATE Obtain ${\bf{a}}^{(t)}_k, \forall k \in \mathcal{K}$ by removing the last element of $\bar{\bf{a}}^{(t)}_k $ and set $t'=0$.
				\REPEAT
				\STATE Update $t' \leftarrow t'+1$.
				\STATE Given ${{\bf{b}}^{(t)}}$, $\{ {\bf{a}}_k^{(t)} \}$, ${\bf{\Theta}}^{(t-1)}$, obtain ${\bf{f}}^{(t')}$ by solving (\ref{p_9}) using CVX with ${\bf{f}}^{(t'-1)}$.
				\UNTIL $t' \ge T_2$ or convergence accuracy reaches $\varepsilon_2$;
				\STATE Set ${\bf{f}}^{(t)}={\bf{f}}^{(t')}$ and $t'=0$.
				\REPEAT
				\STATE Update $t' \leftarrow t'+1$.
				\STATE Given ${{\bf{b}}^{(t)}}$, $\{ {\bf{a}}_k^{(t)} \}$, ${\bf{f}}^{(t)}$, obtain ${{\bf{v}}^{(t')}}$ by solving (\ref{p_11}) using CVX with ${\bf{v}}^{(t'-1)}$.
				\UNTIL $t' \ge T_3$ or convergence accuracy reaches $\varepsilon_3$;
				\STATE Set ${{\bf{v}}^{(t')}}$ =  ${{\bf{v}}^{(t')}}$~\emph{mod}~$2\pi$ .
				\STATE Recover ${\bf{\Theta}}^{(t)}=\text{diag}( {{\bf{v}}^{(t')}} )$ and set $t'=0$.
				\UNTIL $t \ge {T_0}$ or convergence accuracy reaches $\varepsilon_0$;
				\STATE Set $({{\bf{b}}},\{ {\bf{a}}_k \},{\bf{f}},{\bf{\Theta}}) = ({{\bf{b}}^{(t)}},\{ {\bf{a}}_k^{(t)} \},{\bf{f}}^{(t)},{\bf{\Theta}}^{(t)})$.
				\STATE \textbf{Output} the optimized solution $({{\bf{b}}},\{ {\bf{a}}_k \},{\bf{f}},{\bf{\Theta}})$.
		\end{algorithmic}
	\end{algorithm}
	
	\subsection{Algorithm, Complexity and Convergence} 
	\label{algorithm_perfect_CSI}
	Algorithm 1 summarizes the proposed AO-based algorithm, where the four stages described in Sections \ref{receive_computation_beamforming_vector_perfect} to {\ref{phase_shift_matrix_perfect_CSI}} repeat until the convergence accuracy or the maximum iteration number is reached.
	The complexity of Algorithm 1 is $\mathcal{O} (T_{0}+T_{0}T_1(N_t+1)^4+T_{0}{T_{2}{(N_r+1)}^3}+T_{0}{T_{3}M^3})$, where $T_0$ is the number of iterations for AO; $T_1$ is the number of iterations needed to solve $\{{\bf{a}}_k\}$ with DC programming; $T_2$ is the number of iterations needed to solve ${\bf{f}}$ with SCA; and $T_3$ is the number of iterations needed to solve ${\bf{\Theta}}$ with SCA.
	Specifically, the complexity of optimizing $\bf{b}$ in (\ref{optimal_b_perfect_CSI}) is $\mathcal{O} (1)$.
	The complexity of performing DC programming to solve $\{{{\bf{a}}_k}\}$ is $\mathcal{O} (T_1(N_t+1)^4)$~\cite{Boyd2004Convex}.
	As for $\bf{f}$ and $\bf{\Theta}$, we take the interior point method at each iteration of the SCA.
	The complexities of solving $\bf{f}$ and $\bf{\Theta}$ are $\mathcal{O} (T_{2}{(N_r+1)}^3)$ and $\mathcal{O} (T_{3}M^3)$, respectively~\cite{Wanli2021Resource}.
	
	The convergence of Algorithm 1 is briefly demonstrated using the Monotone Bounded theorem~\cite{Ni2021FL}.
	Specifically, the value of the objective funtion (\ref{p_1_MSE}) is non-increasing throughout the AO iterations, because the four subproblems minimize or reduce the objective in an alternating fashion.
	On the other hand, the objective function (\ref{p_1_MSE}) is non-negative and hence lower bounded.
	The convergence of Algorithm 1 is confirmed.
	
	\section{Robust Beamforming and RIS Configuration under Imperfect CSI} \label{solution_imperfect_CSI}
	Under imperfect CSI, a robust design of the beamformers of the devices and BS, and the phase shifts of the RIS is important for the accuracy and integrity of AirFL. 
	This section derives the MSE and the average SINR under imperfect CSI, and accordingly the robust design.

	The superposition signals for model aggregation, ${{\tilde s}_b}$, and local model recovery, ${\tilde s}_f$, are
	%
	\begin{align}
		\label{signal_b_imperfect_CSI}
		{{\tilde s}_b} \!=& {\sum\nolimits_{k = 1}^K {{\bf{b}}^{\rm H} {\widehat {\bf{H}}_k} {{\bf{a}}_k} {s_k}}} \!+ {\sum\nolimits_{k = 1}^K {{\bf{b}}^{\rm H} \Delta {\bf{H}}_k}{{\bf{a}}_k}{s_k}} \!+\! {\bf{b}}^{\rm H}{\bf{n}}, \\
		\label{signal_f_imperfect_CSI}
		{\tilde s}_f \!=& {\sum\nolimits_{k = 1}^K {{\bf{f}}^{\rm H} {\widehat {\bf{H}}_k} {{\bf{a}}_k} {s_k}}} \!+ {\sum\nolimits_{k = 1}^K {{\bf{f}}^{\rm H} \Delta {\bf{H}}_k}{{\bf{a}}_k}{s_k}} \!+\! {\bf{f}}^{\rm H}{\bf{n}}.
	\end{align}

	\begin{lemma}
		\label{lemma_2}
		Under imperfect CSI, the MSE between the detected superposition signal ${\tilde s}_b$ and the desired aggregated AirFL model $s$ is given by
		\begin{align}
			\label{MSE_imperfect}
			{\rm MSE}\left( {\tilde s}_b,s \right)=&{\sum\nolimits_{k = 1}^K {|{{\bf{b}}^{\rm H}}{\widehat {\bf{H}}_k}{{\bf{a}}_k} - 1|}^2}+ {{\sigma}^2_n}{\left\| {\bf{b}} \right\|^2} \notag \\
			&+ {\sum\nolimits_{k = 1}^K {{{\bf{b}}^{\rm H}}{{\bf{J}}_k}{\bf{b}}}},
		\end{align}
		where ${{\bf{J}}_k}$ is defined for notational simplicity, as given by
		\begin{align}
			\label{Jk_imperfect_CSI}
			{{\bf{J}}_k} \buildrel \Delta \over =&( {\sigma _{d,k}^2}{\| {{\bf{a}}_k} \|}^2 + {\sigma _{g}^2}{\| {{\widehat {\bf{H}}_{r,k}}}{{\bf{a}}_k} \|}^2 + M{\sigma _{r,k}^2}{\sigma _{g}^2}{\| {{\bf{a}}_k} \|}^2 ) {{\bf{I}}_{N_r}} \notag \\ 
			&+ {\sigma _{r,k}^2}{\| {{\bf{a}}_k} \|}^2{{\widehat {\bf{G}}}^{\rm H}}{\widehat {\bf{G}}}, \forall k \in \mathcal{K}.
		\end{align}
	\end{lemma}
	
	%
	\begin{IEEEproof}
		Please refer to Appendix \ref{proof_of_lemma_2}.
	\end{IEEEproof}
	
	From (\ref{signal_f_imperfect_CSI}), the noise and interference combined for the $k$-th device, $e_k$, is given by
	%
	\begin{align}
		e_k=&\underbrace{{\sum\nolimits_{k' = k+1}^K {{\bf{f}}^{\rm H} {\widehat {\bf{H}}_{k'}} {{\bf{a}}_{k'}} {s_{k'}}}}}_{\text{interference~from~non-recovered~devices}}+\underbrace{\sum\nolimits_{k' = 1}^K {\bf{f}}^{\rm H} \Delta {\bf{H}}_{k'} {{\bf{a}}_{k'}}{s_{k'}}}_{\text{interference~due~to~imperfect~CSI}} \notag \\ 
		&+ {\bf{f}}^{\rm H}{\bf{n}}, \forall k \in \mathcal{K}.
	\end{align}
	The average interference-plus-noise power $\Sigma_k=\mathbb{E}[e^{\rm H}_k e_k]$ is given by
	%
	\begin{align}
		\Sigma_k \! =&\! \sum\nolimits_{k' = k + 1}^K \mathbb{E}[ {{ | {{{\bf{f}}^{\rm H}}{{\widehat {\bf{H}}}_{k'}}{{\bf{a}}_{k'}}}{s_{k'}} |^2}} ] + \mathbb{E} [ | {\bf{f}}^{\rm H}{\bf{n}} |^2 ] \notag \\
		& + \sum\nolimits_{k' = 1}^K \mathbb{E} [ | {\bf{f}}^{\rm H} ( {\Delta {\bf{H}}_{d,k'}} + {{\widehat {\bf{G}}}^{\rm H}}{\bf{\Theta}}{\Delta {\bf{H}}_{r,k'}}  \notag \\ 
		&+ {{{\Delta {\bf{G}}}^{\rm H}}{{\bf{\Theta}}}{{\widehat {\bf{H}}_{r,k'}}}}+ {{{\Delta {\bf{G}}}^{\rm H}}{{\bf{\Theta}}}{\Delta {\bf{H}}_{r,k'}}} ){{\bf{a}}_{k'}}{s_{k'}} |^2 ]  \notag \\
		=&\! \sum\nolimits_{k' = k + 1}^K \!\! {{\left| {{{\bf{f}}^{\rm H}}{{\widehat {\bf{H}}}_{k'}}{{\bf{a}}_{k'}}} \! \right|}^2} \hspace{-0.2 cm} + \! \sigma _n^2{{\left\| {\bf{f}} \right\|}^2} \!\! + \hspace{-0.1 cm} \sum\nolimits_{k' = 1}^K \!\! {{{\bf{f}}^{\rm H}}{{\bf{J}}_{k'}}{\bf{f}}}, \forall k \in \mathcal{K}.
	\end{align}
	
	%
	\noindent Since the channel estimation errors are agnostic to the BS, the local models are recovered serially in the descending order of the Frobenius norms of the estimated channels of the devices.
	The average SINR of the $k$-th device, i.e., $\tilde{\gamma}_k$, is~\cite{Iimori2019MIMO}
	%
	\begin{equation}
		\label{SINR_imperfect_CSI}
		{\tilde{\gamma}_k} = \frac{{{{\left| {{{\bf{f}}^{\rm H}}{{\widehat {\bf{H}}}_k}{{\bf{a}}_k}} \right|}^2}}}{{\sum\limits_{k' = k + 1}^K {{{\left| {{{\bf{f}}^{\rm H}}{{\widehat {\bf{H}}}_{k'}}{{\bf{a}}_{k'}}} \right|}^2} \!+\! \sum\limits_{k' = 1}^K {{{\bf{f}}^{\rm H}}{{\bf{J}}_{k'}}{\bf{f}}} \!+\! \sigma _n^2{{\left\| {\bf{f}} \right\|}^2}} }},\forall k \in \mathcal{K}.
	\end{equation}
	
	\vspace{-0.1cm}
	\noindent By comparing (\ref{SINR_imperfect_CSI}) with (\ref{SINR_perfect_CSI}), we see that the channel estimation errors cause stronger interference.
	Larger signal power gaps are required for successful SIC.
	(\ref{perfect_SIC_power_difference}) is updated as
	%
	\begin{equation}
		\label{imperfect_SIC_power_difference}
		\hspace{-0.02 cm}{\left| {{{\bf{f}}^{\rm H}}{{\widehat {\bf{H}}_k}}{{\bf{a}}_k}} \! \right|^2} \!\!-\! \sum\nolimits_{k' = k + 1}^K \! {{{\left| {{{\bf{f}}^{\rm H}}{{\widehat {\bf{H}}_{k'}}}{{\bf{a}}_{k'}}} \! \right|}^2}} \!\! \ge \! {{\tilde p}_{\rm gap}},\forall k \in \mathcal{K}\backslash \{ K\},\!\!\!\!
	\end{equation}

	\noindent where ${\tilde p}_{\rm gap}$ denotes the minimum required power gap under imperfect CSI.

	Given (\ref{MSE_imperfect}), (\ref{SINR_imperfect_CSI}) and (\ref{imperfect_SIC_power_difference}), we formulate the problem of interest under imperfect CSI as
	%
	\begin{subequations}
		\begin{eqnarray}
			\label{p_12_MSE}
			&{\hspace{-1.1	cm}}\mathop {\min }\limits_{{\bf{b}},{\bf{f}},{\bf{\Theta}},\atop{\{{{\bf{a}}_k}\}}} {\hspace{-0.3 cm}}& {\sum\nolimits_{k = 1}^K \! {|{{\bf{b}}^{\rm H}}{\widehat {\bf{H}}_k}{{\bf{a}}_k} \!-\! 1|}^2} \!+\! {{\sigma}^2_n}{\left\| {\bf{b}} \right\|^2} \!+\! {\sum\nolimits_{k = 1}^K \! {{{\bf{b}}^{\rm H}}{{\bf{J}}_k}{\bf{b}}}} \\
			\label{QoS_constraints_imperfect_CSI}
			&{\hspace{-1.1 cm}}{\rm s.t.} {\hspace{-0.3 cm}}&{\tilde{\gamma}_k} \ge {\gamma_{\min}}, \forall k \in {\mathcal{K}}, \\
			&{\hspace{-1.1 cm}}{}{\hspace{-0.3 cm}}&\text{(\ref{transmit_power_constraints_perfect_CSI})},~ \text{(\ref{phase_shift_constraints_perfect_CSI})},~ \text{(\ref{imperfect_SIC_power_difference})}. \notag
		\end{eqnarray}
		\label{p_12}
	\end{subequations}
	
	\vspace{-0.6 cm}
	\noindent Problem (\ref{p_12}) is non-convex, and more challenging than problem (\ref{p_1}) because of the non-convexity of (\ref{imperfect_SIC_power_difference}), (\ref{QoS_constraints_imperfect_CSI}) and the new terms involving $\{{\bf{J}}_k\}$.
	
	\subsection{Robust Design under Imperfect CSI}
	As done in Section~\ref{solution_perfect_CSI}, we decompose problem (\ref{p_12}) into four subproblems, and employ the AO method to solve the problem under imperfect CSI. 
	%
	
	\subsubsection{Receive Beamformer for Model Aggregation}
	Given fixed $\{{\bf{a}}_k\}$, $\bf{f}$ and $\bf{\Theta}$, problem (\ref{p_12}) is reduced to an unconstrained subproblem regarding $\bf{b}$, as given by
	%
	\begin{align}
		{\hspace{-0.25 cm}}\mathop {\min }\limits_{{\bf{b}}}\hspace{+0.1 cm} & {{\bf{b}}^{\rm{H}}}\hspace{-0.15 cm}\left[ {\sum\limits_{k = 1}^K \! {\left( {{\tilde{\bf{H}}}_{a,k}}\! +\! {\bf{J}}_k \right)}\hspace{-0.1 cm} +\! \sigma _n^2{{\bf{I}}_{{N_r}}}}\hspace{-0.12 cm} \right]\hspace{-0.1 cm}{\bf{b}} \!-\! 2{\mathop{\rm Re}\nolimits}\! \left\{\hspace{-0.1 cm} {{{\bf{b}}^{\rm{H}}}\!\sum\limits_{k = 1}^K \! {{{\widehat {\bf{H}}}_k}{{\bf{a}}_k}} }\hspace{-0.1 cm} \right\} \!\!
		\label{p_13}
	\end{align}
	where ${{{\tilde{\bf{H}}}_{a,k}}} = {\widehat {\bf{H}}_k}{{\bf{a}}_k}{{\bf{a}}_k^{\rm H}}{\widehat {\bf{H}}_k^{\rm H}}$.
	Problem (\ref{p_13}) is convex given the positive semidefinite matrices $\{{{\tilde{\bf{H}}}_{a,k}} + {\bf{J}}_k\}$.
	As done in (\ref{optimal_b_perfect_CSI}), we find $\bf{b}$ in closed-form by the MMSE criterion, as given by
	\vspace{-0.2 cm}
	\begin{equation}
		\label{optimal_b_imperfect_CSI}
		{\bf{b}} \! = \!\! {\left[ {\sum\nolimits_{k = 1}^K \! {\left( {{\tilde{\bf{H}}}_{a,k}} \!+\! {\bf{J}}_k \right)}  \!+\! \sigma _n^2{{\bf{I}}_{{N_r}}}} \right]}^{-1} \!\! \left(\sum\nolimits_{k = 1}^K \! {{{\widehat {\bf{H}}}_k}{{\bf{a}}_k}} \right).
	\end{equation}
	%
	
	\subsubsection{Transmit Beamformers}
	Given fixed $\bf{b}$, $\bf{f}$ and $\bf{\Theta}$, by replacing $\{{\bf{a}}_k\}$ with $\{{\bf{A}}_k\}$ as in Section \ref{transmit_beamforming_vectors_perfect}, problem (\ref{p_12}) is reduced to an SDP problem concerning $\{{\bf{A}}_k\}$,
	\vspace{-0.2 cm}
	\begin{subequations}
		\begin{eqnarray}
			&{\hspace{-0.4 cm}}\mathop {\min }\limits_{\{{{\bf{A}}_k}\}} & \sum\nolimits_{k = 1}^K {\text{tr}({{\tilde{\bf{Z}}}_{0,k}}{{\bf{A}}_k})}  \\
			\label{SDR_constraints_1_imperfect}
			&{\hspace{-0.4 cm}}{\rm s.t.}&-\text{tr}({\tilde {\bf{Z}}_{2,k}}{{\bf{A}}_k})\!\! +\! {\gamma_{\min}}{\sum\nolimits_{k' = 1}^K{\hspace{-0.1 cm}} {\text{tr}({{\tilde{\bf{Z}}}_{1,k'}}{{\bf{A}}_{k'}})}}\!\! +\! {{\gamma_{\min}}{\sigma_n^2}\|{\bf{f}}\|^2}\notag \\
			&{}&+ {\gamma_{\min}} {\sum\nolimits_{k' = k + 1}^K{\hspace{-0.2 cm}} {\text{tr}({\tilde{\bf{Z}}_{2,k'}}{{\bf{A}}_{k'}})}}\! \le \! 0, \forall k \in {\mathcal{K}}, \\
			\label{SDR_constraints_2_imperfect}
			&{\hspace{-0.4 cm}}&-\text{tr}({\tilde{\bf{Z}}_{2,k}}{{\bf{A}}_k})\! +\! {\sum\nolimits_{k' = k + 1}^K {\text{tr}({\tilde{\bf{Z}}_{2,k'}}{{\bf{A}}_{k'}})}} \notag \\
			&{}&+{{\tilde p}_{\rm gap}} \le 0, \forall k \in \mathcal{K}\backslash \{ K\}, \\
			&{}&\text{(\ref{SDR_constraints_1_perfect})},~\text{(\ref{SDR_constraints_4_perfect})},~\text{(\ref{SDR_constraints_5_perfect})},~\text{(\ref{SDR_constraints_6_perfect})}, \notag
		\end{eqnarray}
		\label{p_14}
	\end{subequations}
	
	\vspace{-0.4 cm}
	\noindent where
	%
	%
	\begin{align}
		&{\hspace{-0.78 cm}}{\bf{D}}_{b,k} \buildrel \Delta \over = ( {\sigma _{d,k}^2}{\| {\bf{b}} \|}^2  +  M{\sigma _{r,k}^2}{\sigma _{g}^2}{\| {\bf{b}} \|}^2  +  {\sigma _{r,k}^2}{\| {{\widehat {\bf{G}}}}{\bf{b}} \|}^2 ) {{\bf{I}}_{N_t}} \notag \\
		&\quad+  {\sigma _{g}^2}{\| {{\bf{b}}} \|}^2{{\widehat {\bf{H}}_{r,k}}^{\rm H}}{\widehat {\bf{H}}_{r,k}}, \forall k \in \mathcal{K}, \\
		&{\hspace{-0.78 cm}}{\bf{D}}_{f,k} \buildrel \Delta \over = ( {\sigma _{d,k}^2}{\| {\bf{f}} \|}^2 + M{\sigma _{r,k}^2}{\sigma _{g}^2}{\| {\bf{f}} \|}^2 + {\sigma _{r,k}^2}{\| {{\widehat {\bf{G}}}}{\bf{f}} \|}^2 ) {{\bf{I}}_{N_t}} \notag \\
		&\quad+ {\sigma _{g}^2}{\| {{\bf{f}}} \|}^2{{\widehat {\bf{H}}_{r,k}}^{\rm H}}{\widehat {\bf{H}}_{r,k}}, \forall k \in \mathcal{K},
	\end{align}
	\vspace{-0.6 cm}
	\begin{align}
		&{{\tilde{\bf{Z}}}_{0,k}} \! \buildrel \Delta \over =  {\hspace{-0.2cm}} \left[ {\hspace{-0.2cm}} {\begin{array}{*{20}{c}} {{\widehat {\bf{H}}^{\rm H}_k}{\bf{b}}{{\bf{b}}^{\rm H}}{\widehat {\bf{H}}_k}\!+\!{{\bf{D}}_{b,k}}}{\hspace{-0.2cm}}&{{ - \widehat {\bf{H}}_k^{\rm{H}}{\bf{b}}}}\\
				{- {{\bf{b}}^{\rm{H}}}{\widehat { {\bf{H}}}_k}}{\hspace{-0.2cm}}&0\end{array}} {\hspace{-0.2cm}} \right]\!,
		{{\tilde{\bf{Z}}}_{1,k}} \! \buildrel \Delta \over = {\hspace{-0.2cm}} \left[ {\hspace{-0.2cm}} {\begin{array}{*{20}{c}} {{{\bf{D}}_{f,k}}}{\hspace{-0.2cm}}&{{{\bf{0}}_{{N_t} \times 1}}}\\
				{{\bf{0}}_{{N_t} \times 1}^{\rm{H}}}{\hspace{-0.2cm}}&0\end{array}} {\hspace{-0.2cm}} \right]\!, \notag \\
		&{{\tilde{\bf{Z}}}_{2,k}} \! \buildrel \Delta \over = {\hspace{-0.2cm}} \left[ {\hspace{-0.2cm}} {\begin{array}{*{20}{c}} {{\widehat {\bf{H}}^{\rm H}_k}{\bf{f}}{{\bf{f}}^{\rm H}}{\widehat {\bf{H}}_k}}{\hspace{-0.2cm}}&{{{\bf{0}}_{{N_t} \times 1}}}\\
				{{\bf{0}}_{{N_t} \times 1}^{\rm{H}}}{\hspace{-0.2cm}}&0 \end{array}} {\hspace{-0.2cm}} \right]\!,\forall k \! \in \! \mathcal{K}.
	\end{align}
	
	\vspace{-0.2 cm}
	\noindent Referring to (\ref{equavalent_rank_constraints})--(\ref{p_6}), we invoke DC programming to convexify the non-convex problem (\ref{p_14}):
	%
	%
	\begin{eqnarray}
		\label{p_15}
		{\hspace{-0.5 cm}}&\mathop {\min }\limits_{\{{{\bf{A}}_k}\}} & \sum\nolimits_{k = 1}^K \{\text{tr}(({{\tilde{\bf{Z}}}_{0,k}}+\alpha {\bf{I}}_{N_t}){{\bf{A}}_k}) - \alpha \langle {{\bf{\dot A}}}^{(t)}_k , {\bf{A}}_k \rangle\} \\
		{\hspace{-0.5 cm}}&{\rm s.t.}&\text{(\ref{SDR_constraints_1_perfect})},~\text{(\ref{SDR_constraints_4_perfect})},~\text{(\ref{SDR_constraints_5_perfect})},~\text{(\ref{SDR_constraints_1_imperfect})},~\text{(\ref{SDR_constraints_2_imperfect})}, \notag
	\end{eqnarray}
	
	\vspace{-0.2 cm}
	\noindent which can be solved using CVX.
	$\bar{\bf{a}}_k$ is obtained from the rank-one matrices, $\{{\bf{A}}_k\}$, by eigenvalue decomposition. The solution of ${\bf{a}}_k$ is obtained by removing the last element of $\bar{\bf{a}}_k$, $\bar{\bf{a}}_k=[{\bf{a}}^{\rm H}_k,u^{\rm H}_k]^{\rm H}$ with $u^2_k=1$.
	
	\subsubsection{Receive Beamformer for Local Model Recovery}
	Given fixed $\bf{b}$, $\{{\bf{a}}_k\}$ and $\bf{\Theta}$, problem (\ref{p_12}) is reduced to a feasibility problem w.r.t. $\bf{f}$, as given by
	\vspace{-0.2 cm}
	\begin{subequations}
		\begin{eqnarray}
			&\mathop {\text{find} }\limits_{{\bf{f}}} & {\bf{f}} \\
			&{\rm s.t.}& {\bf{f}}^{\rm H}{\tilde {\bf{B}}_{1,k}}{\bf{f}} \le 0 , \forall k \in {\mathcal{K}}, \\
			&{}&{\bf{f}}^{\rm H}{\tilde {\bf{B}}_{2,k}}{\bf{f}} + {{{\tilde p}_{\rm gap}}} \le 0 , \forall k \in {\mathcal{K}}\backslash \{ K\},
		\end{eqnarray}
		\label{p_16}
	\end{subequations}
	
	\vspace{-0.4 cm}
	\noindent where ${\tilde {\bf{B}}_{1,k}} \buildrel \Delta \over = {\gamma _{\min }}\sum\nolimits_{k' = k + 1}^K {{{\tilde {\bf{H}}}_{a,k'}}} - {{\tilde {\bf{H}}}_{a,k}} + {\gamma _{\min }}\sigma _n^2{{\bf{I}}_{{N_r}}} + {\gamma _{\min}}\sum\nolimits_{k'=1}^K {{\bf{J}}_{k'}}, \forall k \in \mathcal{K}$ and ${\tilde{\bf{B}}_{2,k}} \buildrel \Delta \over = \sum\nolimits_{k' = k + 1}^K {{{\tilde {\bf{H}}}_{a,k'}}} - {{\tilde {\bf{H}}}_{a,k}}, \forall k \in \mathcal{K}\backslash \{ K\}$.
	As done in Section \ref{receive_decoding_beamforming_perfect}, we transform problem (\ref{p_16}) to minimize $\beta \in \mathbb{R}$, as given by
	\vspace{-0.2 cm}
	\begin{subequations}
		\begin{eqnarray}
			&\mathop {\min }\limits_{{\bf{f}},{\beta \le 0}} & {\beta} \\
			&{\rm s.t.}& {\bf{f}}^{\rm H}{\tilde {\bf{B}}_{1,k}}{\bf{f}} \le {\beta} , \forall k \in {\mathcal{K}}, \\
			&{}&{\bf{f}}^{\rm H}{\tilde {\bf{B}}_{2,k}}{\bf{f}} + {{{\tilde p}_{gap}}} \le {\beta} , \forall k \in {\mathcal{K}}\backslash \{ K\},
		\end{eqnarray}
		\label{p_17}
	\end{subequations}
	
	\vspace{-0.6 cm}
	\noindent which is non-convex due to indefinite matrices ${\tilde {\bf{B}}_{1,k}}$ and ${\tilde {\bf{B}}_{2,k}}$.
	We apply SCA to solve problem (\ref{p_17}), where one surrogate function $\tilde{g}_{1,k}\left(\mathbf{f} | \mathbf{f}^{(t)}\right)$ is obtained by replacing $\mathbf{B}_{1,k}$ with ${\tilde {\bf{B}}_{1,k}}$ in (\ref{surrogate_Ak}) and the other surrogate function $\tilde {g}_{2,k}\left(\mathbf{f} | \mathbf{f}^{(t)}\right)$ is obtained by replacing $\mathbf{B}_{2,k}$ with ${\tilde {\bf{B}}_{2,k}}$ in (\ref{surrogate_Bk}).

	Problem (\ref{p_17}) is approximated to a sequence of convex problems w.r.t. $\{\bf{f}, \beta\}$, i.e.,
	\vspace{-0.2 cm}
	\begin{subequations}
		\begin{eqnarray}
			&\mathop {\min }\limits_{{\bf{f}},{\beta\le 0}} & {\beta } \\
			&{\rm s.t.}& {\tilde g_{1,k}}\left( {\bf{f}} | {{\bf{f}}^{(t)}} \right) \le {\beta} , \forall k \in {\mathcal{K}}, \\
			&{}&{\tilde g_{2,k}}\left( {\bf{f}} | {{\bf{f}}^{(t)}} \right) + {{{\tilde p}_{gap}}} \le {\beta} , \forall k \in {\mathcal{K}}\backslash \{ K\},
		\end{eqnarray}
		\label{p_18}
	\end{subequations}
	
	\vspace{-0.4 cm}
	\noindent which can be solved using CVX toolkits.
	
	\begin{algorithm}[t]
		\caption{Proposed Algorithm for Imperfect CSI Case}
		\label{algorithm_2}
		\begin{algorithmic}[1]
			\renewcommand{\algorithmicrequire}{\textbf{Initialize}}
			\renewcommand{\algorithmicensure}{\textbf{Output}}
			\STATE \textbf{Initialize} a feasible solution $({{\bf{b}}^{(0)}},\{ {\bf{a}}_k^{(0)} \},{\bf{f}}^{(0)},{\bf{\Theta}}^{(0)})$, the maximum iteration numbers $T_0$, $T_1$, $T_2$, $T_3$, the convergence accuracies $\varepsilon_0$, $\varepsilon_1$, $\varepsilon_2$, $\varepsilon_3$, and set $t=0$, $t'=0$.
			\REPEAT
			\STATE Update $t \leftarrow t+1$.
			\STATE Given $\{ {\bf{a}}_k^{(t-1)} \}$, ${\bf{f}}^{(t-1)}$, ${\bf{\Theta}}^{(t-1)}$, calculate ${{\bf{b}}^{(t)}}$ via (\ref{optimal_b_imperfect_CSI}).
			\REPEAT
			\STATE Update $t' \leftarrow t'+1$.
			\STATE Calculate ${{\bf{\dot A}}}^{(t'-1)}_k={\bf{u}}^{(t'-1)}_k({\bf{u}}^{(t'-1)}_k)^{\rm H},\forall k \in \mathcal{K}$.
			\STATE Given ${{\bf{b}}^{(t)}}$, ${\bf{f}}^{(t-1)}$, ${\bf{\Theta}}^{(t-1)}$, obtain $\{{\bf{A}}^{(t')}_k\}$ by solving (\ref{p_15}) using CVX.
			\UNTIL $t' \ge T_1$ or $\text{tr}({\bf{A}}^{(t')}_k)-\|{\bf{A}}^{(t')}_k\|_2 \le \varepsilon_1, \forall k \in \mathcal{K}$;
			\STATE Recover $\bar{\bf{a}}^{(t)}_k = \sqrt {{\lambda _{{\bf{A}}^{(t')}_k}}} {{\bf{p}}_{k}}, \forall k \in \mathcal{K}$.
			\STATE Obtain ${\bf{a}}^{(t)}_k, \forall k \in \mathcal{K}$ by removing the last element of $\bar{\bf{a}}^{(t)}_k $ and set $t'=0$.
			\REPEAT
			\STATE Update $t' \leftarrow t'+1$.
			\STATE Given ${{\bf{b}}^{(t)}}$, $\{ {\bf{a}}_k^{(t)} \}$, ${\bf{\Theta}}^{(t-1)}$, obtain ${\bf{f}}^{(t')}$ using CVX with ${\bf{f}}^{(t'-1)}$.
			\UNTIL $t' \ge T_2$ or convergence accuracy reaches $\varepsilon_2$;
			\STATE Set ${\bf{f}}^{(t)}={\bf{f}}^{(t')}$ and $t'=0$.
			\REPEAT
			\STATE Update $t' \leftarrow t'+1$.
			\STATE Given ${{\bf{b}}^{(t)}}$, $\{ {\bf{a}}_k^{(t)} \}$, ${\bf{f}}^{(t)}$, obtain ${{\bf{v}}^{(t')}}$ using CVX with ${\bf{v}}^{(t'-1)}$.
			\UNTIL $t' \ge T_3$ or convergence accuracy reaches $\varepsilon_3$;
			\STATE Set ${{\bf{v}}^{(t')}}$ = ${{\bf{v}}^{(t')}}$~\emph{mod}~$2\pi$.
			\STATE Recover ${\bf{\Theta}}^{(t)}=\text{diag}( {{\bf{v}}^{(t')}} )$ and set $t'=0$.
			\UNTIL $t \ge {T_0}$ or convergence accuracy reaches $\varepsilon_0$;
			\STATE Set $({{\bf{b}}},\{ {\bf{a}}_k \},{\bf{f}},{\bf{\Theta}}) = ({{\bf{b}}^{(t)}},\{ {\bf{a}}_k^{(t)} \},{\bf{f}}^{(t)},{\bf{\Theta}}^{(t)})$.
			\STATE \textbf{Output} the optimized solution $({{\bf{b}}},\{ {\bf{a}}_k \},{\bf{f}},{\bf{\Theta}})$.
		\end{algorithmic}
	\end{algorithm}

	\subsubsection{Phase Shift Matrix}
	Given fixed $\bf{b}$, $\{{\bf{a}}_k\}$ and $\bf{f}$, problem (\ref{p_12}) reduces to a subproblem of $\bf{\Theta}$.
	Since ${\bf{J}}_k$ is independent of $\bf{\Theta}$, the subproblem can be rewritten based on Section~\ref{phase_shift_matrix_perfect_CSI}, i.e.,
	%
	\begin{subequations}
		\begin{eqnarray}
			&\mathop {\min }\limits_{ {\bf{v}} } & {{\tilde h_0}\left( {\bf{v}} \right)} \\
			&{\rm s.t.}& {\tilde h_{1,k}}\left( {\bf{v}} \right) \le 0 , \forall k \in {\mathcal{K}}, \\
			&{}&{\tilde h_{2,k}}\left( {\bf{v}} \right) \le 0 , \forall k \in {\mathcal{K}}\backslash \{ K\},
		\end{eqnarray}
		\label{p_19}
	\end{subequations}
	
	\vspace{-0.4 cm}
	\noindent where ${{\tilde h_0}\left( {\bf{v}} \right)}$ is obtained by replacing ${ {\bf{H}}_{d,k}}$, ${ {\bf{H}}_{r,k}}$ and ${ {\bf{G}}}$ in ${{h_0}\left( {\bf{v}} \right)}$ with ${\widehat {\bf{H}}_{d,k}}$, ${\widehat {\bf{H}}_{r,k}}$ and ${\widehat {\bf{G}}}$, respectively; ${\tilde h_{1,k}}\left( {\bf{v}} \right)$ is obtained by replacing ${ {\bf{H}}_{d,k}}$, ${ {\bf{H}}_{r,k}}$, ${ {\bf{G}}}$ and $C_{1,k}$ in ${h_{1,k}}\left( {\bf{v}} \right)$ with ${\widehat {\bf{H}}_{d,k}}$, ${\widehat {\bf{H}}_{r,k}}$, ${\widehat {\bf{G}}}$ and ${\tilde C}_{1,k}$, respectively; and ${\tilde h_{2,k}}\left( {\bf{v}} \right)$ is obtained by replacing ${ {\bf{H}}_{d,k}}$, ${ {\bf{H}}_{r,k}}$, ${ {\bf{G}}}$ and $C_{2,k}$ in ${h_{2,k}}\left( {\bf{v}} \right)$ with ${\widehat {\bf{H}}_{d,k}}$, ${\widehat {\bf{H}}_{r,k}}$, ${\widehat {\bf{G}}}$ and ${\tilde C}_{2,k}$, respectively.
	${\tilde C}_{1,k}$ and ${\tilde C}_{2,k}$ are
	\begin{align}
		{\tilde C_{1,k}} \buildrel \Delta \over =& {\gamma _{\min }}\sum\nolimits_{k' = k + 1}^K \! {|{{\bf{f}}^{\rm{H}}}{{\widehat {\bf{H}}}_{d,k'}}{{\bf{a}}_{k'}}|^2} \!\! - \!\! |{{\bf{f}}^{\rm{H}}}{\widehat {\bf{H}}_{d,k}}{{\bf{a}}_k}|^2 \notag \\ 
		&+ \! {\gamma _{\min }}\sigma _n^2{\left\| {\bf{f}} \right\|^2} \!+ \! {\gamma _{\min}}\sum\nolimits_{k'=1}^K \! {{\bf{f}}^{\rm H}{\bf{J}}_{k'}{\bf{f}}}, \forall k \in \mathcal{K}, \\
		{\tilde C_{2,k}} \buildrel \Delta \over =& \sum\nolimits_{k' = k + 1}^K {|{{\bf{f}}^{\rm{H}}}{{\widehat {\bf{H}}}_{d,k'}}{{\bf{a}}_{k'}}|^2} \! - \! |{{\bf{f}}^{\rm{H}}}{\widehat {\bf{H}}_{d,k}}{{\bf{a}}_k}|^2 \notag \\ 
		&- \! {{\tilde p}_{\rm gap}}, \forall k \in \mathcal{K}\backslash \{K\}.
	\end{align}

	\noindent The other notations are consistent with those in Table~\ref{table_I}.

	Since problem (\ref{p_19}) is non-convex, we find the phase shift matrix by resorting to the SCA again. 
	The surrogate functions are constructed as
	\vspace{-0.2 cm}
	\begin{align}
		\label{surrogate_hn_bar}
		{\bar h_l}\left( {\bf{v}} | {\bf{v}}^{(t)} \right) =& {\tilde h_l}\left( {\bf{v}}^{(t)} \right)+ \nabla {\tilde h_l}{\left( {{{\bf{v}}^{(t)}}} \right)^{\rm{T}}}\left( {{\bf{v}} - {{\bf{v}}^{(t)}}} \right) \notag \\ 
		&+ \frac{{{\tilde \xi _l}}}{2}{\left\| {{\bf{v}} - {{\bf{v}}^{(t)}}} \right\|^2}, \forall l \in \mathcal{L},
	\end{align} 
	where $\{{\tilde \xi _l}\}$ can still be determined based on Lemma \ref{lemma_1} since $\{{\bf{F}}_l\}$ is independent of the channel estimation errors.
	As a result, problem (\ref{p_19}) can be convexified as
	\vspace{-0.2 cm}
	\begin{subequations}
		\begin{eqnarray}
			&\mathop {\min }\limits_{ {\bf{v}} } & {{\bar h_0}\left( {\bf{v}} | {\bf{v}}^{(t)} \right)} \\
			&{\rm s.t.}& {\bar h_{1,k}}\left( {\bf{v}} | {\bf{v}}^{(t)} \right) \le 0 , \forall k \in {\mathcal{K}}, \\
			&{}&{\bar h_{2,k}}\left( {\bf{v}} | {\bf{v}}^{(t)} \right) \le 0 , \forall k \in {\mathcal{K}}\backslash \{ K\}.
		\end{eqnarray}
		\label{p_20}
	\end{subequations}
	
	\vspace{-0.4 cm}
	\noindent which can be solved using CVX toolkits.
	
	Algorithm 2 summarizes the robust design of the beamformers and  RIS under imperfect CSI.
	Given the same structure of Algorithms 1 and 2, the complexity and convergence of Algorithm 2 can be analyzed in the same way as those of Algorithm 1 and suppressed for brevity.
	The proposed algorithms, i.e., Algorithms 1 and 2, can be readily applied in the absence of the RIS by skipping the part solving the phase shifts, i.e., solving subproblems (\ref{p_10}) or (\ref{p_19}).

	\section{Numerical and Experimental Results} \label{numerical_results}

	In this section, we assess the proposed algorithms by conducting extensive experiments under a setting consistent with~\cite{Gong2020Toward} in which general RIS-aided systems were studied.
	Our experiments involve synthetic wireless channels based on empirical channel fading models for AirComp, and the MNIST or Fashion-MNIST datasets. Both the MNIST and Fashion-MNIST datasets have been extensively used to examine the classification performance of machine learning algorithms~\cite{Cao2022Transmission,Zhang2021Gradient,Zhou2020Privacy,Xiao2017Fashion}. The empirical channel fading models, i.e., the Rician or Rayleigh fading, have been extensively considered in numerical validations of wireless systems~\cite{Yang2020Federated,Ni2021FL,Zhao2021Exploiting}. The Gaussian estimation errors are considered under imperfect CSI, as in~\cite{Ang2019Robust,Mohammad2021Blind,Rong2011Robust,Nosrat2011MIMO}. Nevertheless, the proposed algorithms are general, do not rely on the a-priori knowledge of the channel models, and are not restricted to particular channel models. The algorithms can be applied under other channel fading types.

	In the simulation, the devices are uniformly randomly distributed in a square area of $100 \times 100$ m$^2$ on the ground.
	The three-dimensional coordinates of the BS and RIS are $(0,0,25)$ m and $(20,20,20)$ m, respectively.
	The path loss is $L={C_0}({d}/{D_0})^{-\nu}$, where $C_0$ denotes the path loss at the reference distance $D_0=1$ m and $C_0 = 0~\text{dBm}$ by default, $d$ is the distance between the transmitter and receiver, and $\nu$ is the path loss exponent.
	The path loss exponents from the devices to the BS, from the devices to the RIS, and from the RIS to the BS are set to $\nu_{db}=3.2$, $\nu_{dr}=2.6$, and $\nu_{rb}=2.2$, respectively.
	We consider Rician fading between the devices and RIS, and between the RIS and BS with the Rician factors of $\kappa_{dr}=10$ and $\kappa_{rb}=10$, respectively.
	Two scenarios are considered between the devices and BS.
	In the first scenario, we consider the Rician fading with Rician factor $\kappa_{db}=2$.
	In the second scenario, we consider the Rayleigh fading to emulate the situation where the devices are located in blind spots, i.e., the LoS paths from the devices to the BS are blocked.
	The channel estimation errors are measured by the NMSE $\iota = \mathbb{E}[\|{\bf{H}}-{\widehat {\bf{H}}}\|_F^2]/\mathbb{E}[\|{\widehat {\bf{H}}}\|_F^2]$, over all links~\cite{Zeng2020joint}.
	$\iota=0$ indicates perfect CSI.
	Unless otherwise specified, other parameters are listed in Table~\ref{table_II}.
	
	\begin{table}[t]
		\caption{Simulation Parameters}
		\label{table_II}
		\centering
		{\small
			\begin{tabular}{|l|l|}
				\hline
				\textbf{Parameter} & \textbf{Value} \\ \cline{1-2}
				
				Number of devices & $K=3$ \\ \cline{1-2}
				
				\tabincell{l}{Number of antennas at the BS} & \tabincell{l}{$N_r=16$} \\ \cline{1-2}
				
				\tabincell{l}{Number of antennas at the devices} & \tabincell{l}{$N_t=2$} \\ \cline{1-2}
				
				\tabincell{l}{Number of reflecting elements of the RIS} & $M=40$ \\ \cline{1-2}
				
				\tabincell{l}{Required minimum SINR} & ${\gamma_{\min}}=26.17$~\text{dBm} \\ \cline{1-2}
				
				\tabincell{l}{Maximum transmit power of devices} & $P_{\max}=30$~\text{dBm} \\ \cline{1-2}
				
				\tabincell{l}{Minimum power gap} & \tabincell{c}{ ${\hat p}_{\rm gap}=10$~\text{dBm}, \\ ${\tilde p}_{\rm gap}=17$~\text{dBm} } \\ \cline{1-2}
				
				\tabincell{l}{NMSE of channel estimation error} & $\iota=0.1$ \\ \cline{1-2}
				
				\tabincell{l}{Noise power} & ${{{\sigma}^2_n}}=-80$~\text{dBm} \\ \cline{1-2}
				
				\tabincell{l}{Penalty factor} & $\alpha=1$ \\ \cline{1-2}
				\hline
		\end{tabular}}
		\vspace{-0.4 cm}
	\end{table}

	The devices employ AirFL for classification tasks on the MNIST or Fashion-MNIST dataset~\cite{Xiao2017Fashion}.
	The training data is i.i.d. among the devices. Specifically, each device has $2,000$ non-repetitive $28\times28$ gray-scale images. For each device, there are $10$ categories of images, $200$ images per category.
	Each device trains a fully-connected MLP with a $20$-neuron hidden layer.
	The mini-batch gradient descent is adopted to train the MLP with the learning rate and mini-batch size of $0.01$ and $16$, respectively.
	The devices upload their local models to the BS for aggregation after each training epoch.
	The effectiveness of the aggregated global model is measured with the classification accuracy of the MNIST or Fashion-MNIST test sets.
	
	\begin{figure}[!t]
		\centering
		\begin{minipage}[t]{0.5 \textwidth}
			\centering
			\subfigure[Accuracy on MNIST dataset with the Rician fading between the devices and the BS.]{
					\includegraphics[width=3.5 in]{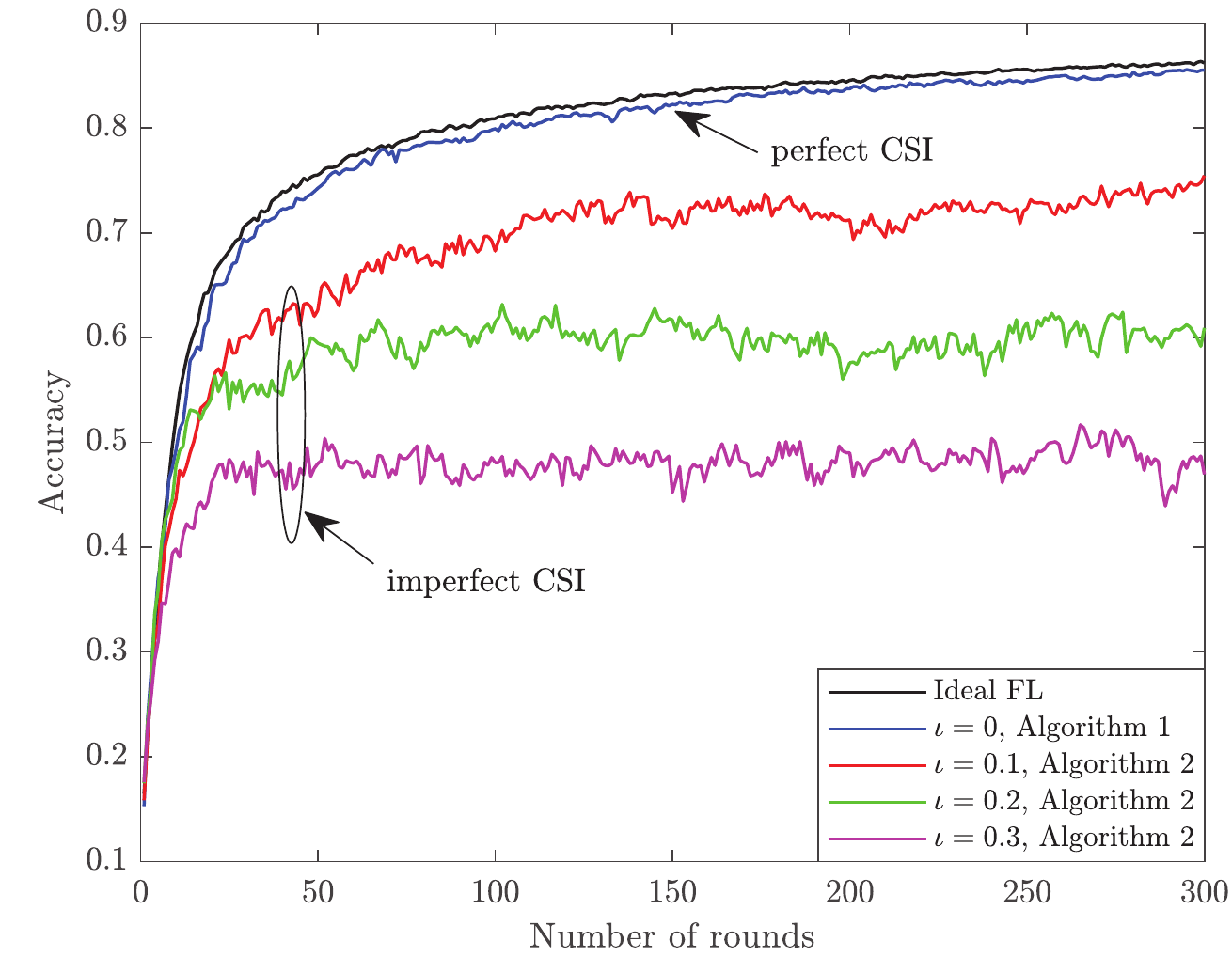}} \\
			\subfigure[Accuracy on Fashion-MNIST dataset with the Rician fading between the devices and the BS.]{
					\includegraphics[width=3.5 in]{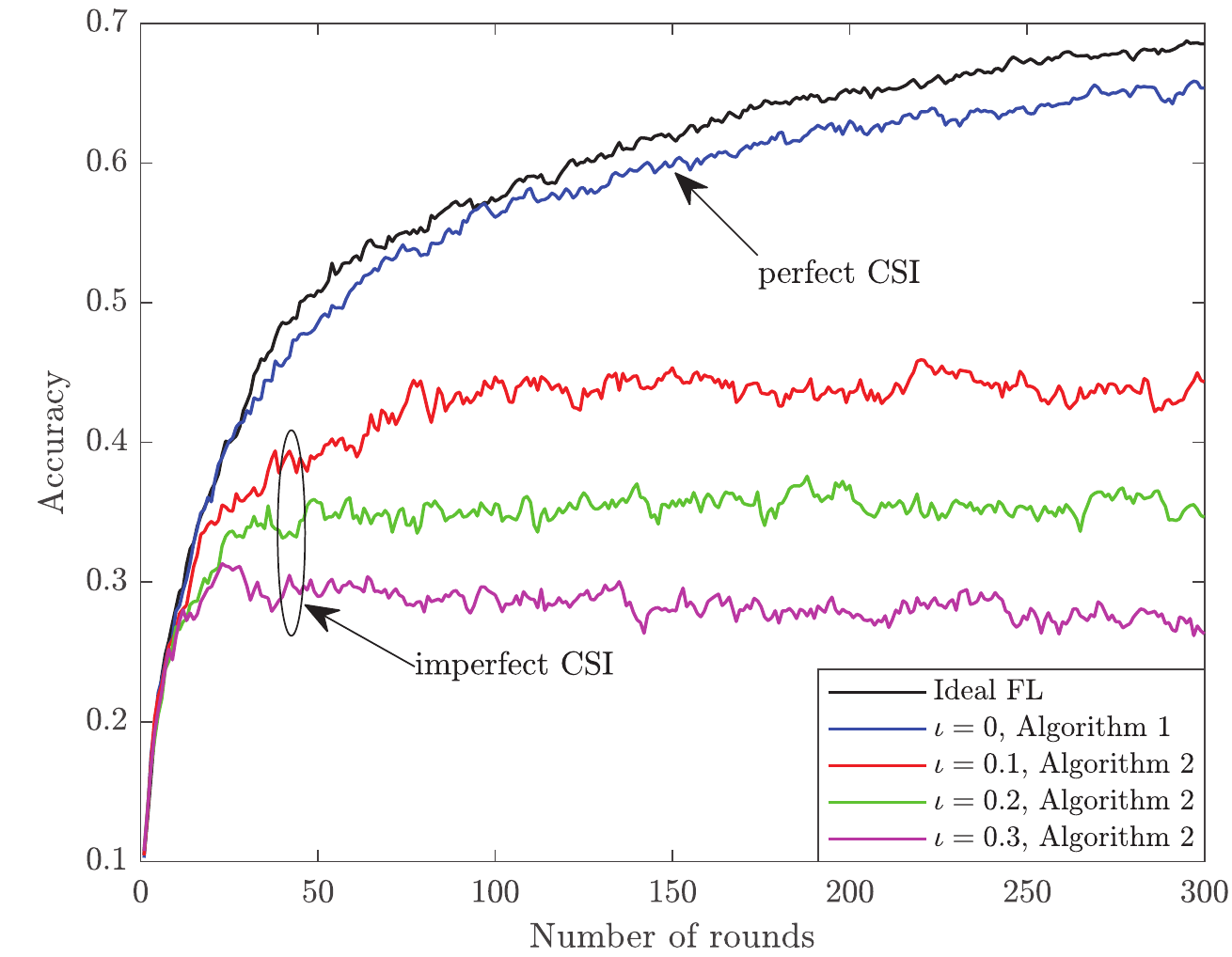}}
			\caption{Accuracy versus the number of rounds, where there are $2000$ training images per device, ${\gamma_{\min}}=26.17$ dBm, $K=3$, $N_t=2$, $N_r=16$, and $M=40$.}
			\label{accuracy_vs_rounds}
		\end{minipage}
		\vspace{-0.4 cm}
	\end{figure}
	\begin{figure}[!t]
		\begin{minipage}[t]{0.5 \textwidth}
			\centering
			\subfigure[Accuracy on MNIST dataset versus $\gamma_{\min}$ with the Rician fading between the devices and the BS.]{
					\includegraphics[width=3.5 in]{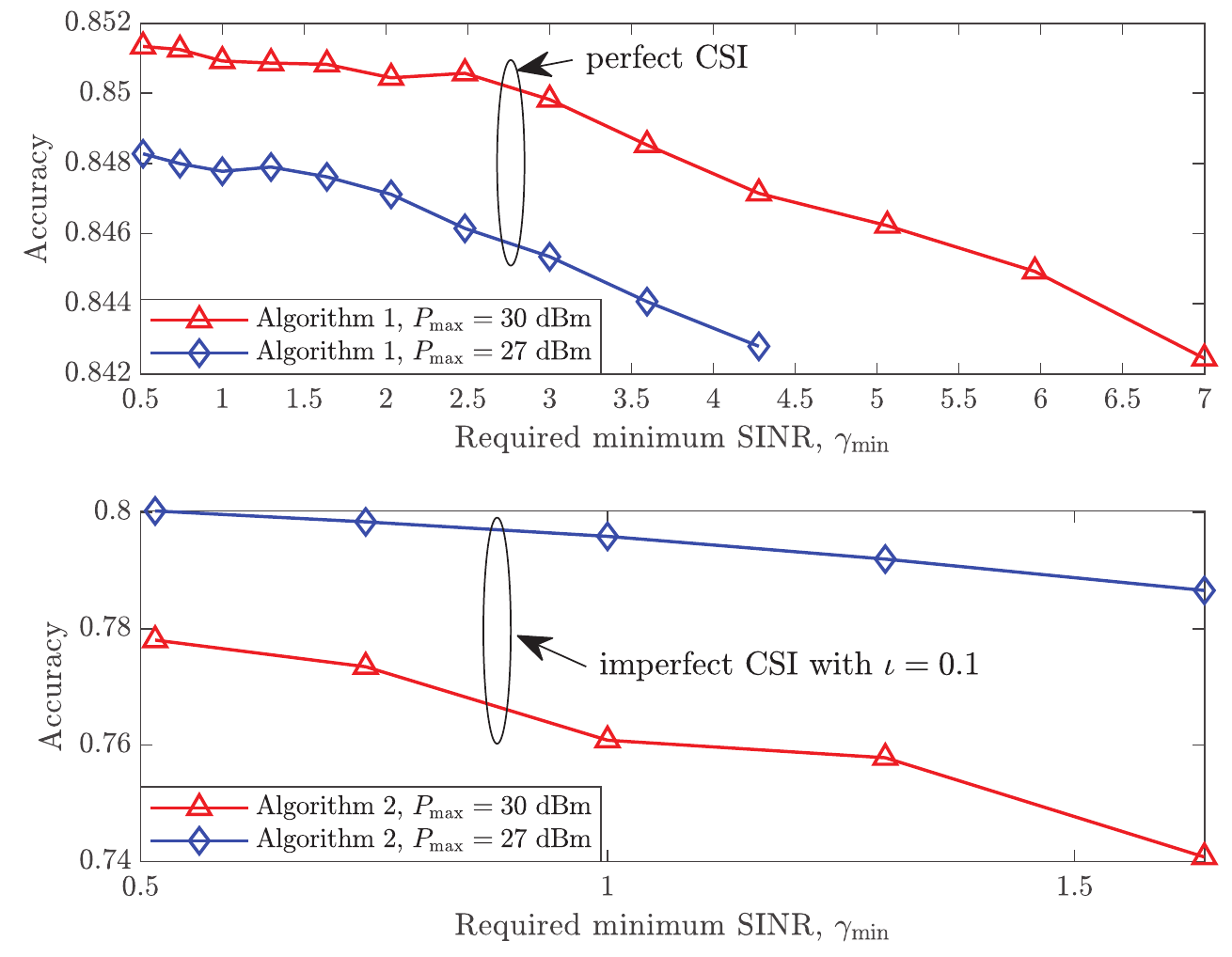}} \\
			\subfigure[Accuracy on Fashion-MNIST dataset versus $\gamma_{\min}$ with the Rician fading between the devices and the BS.]{
					\includegraphics[width=3.5 in]{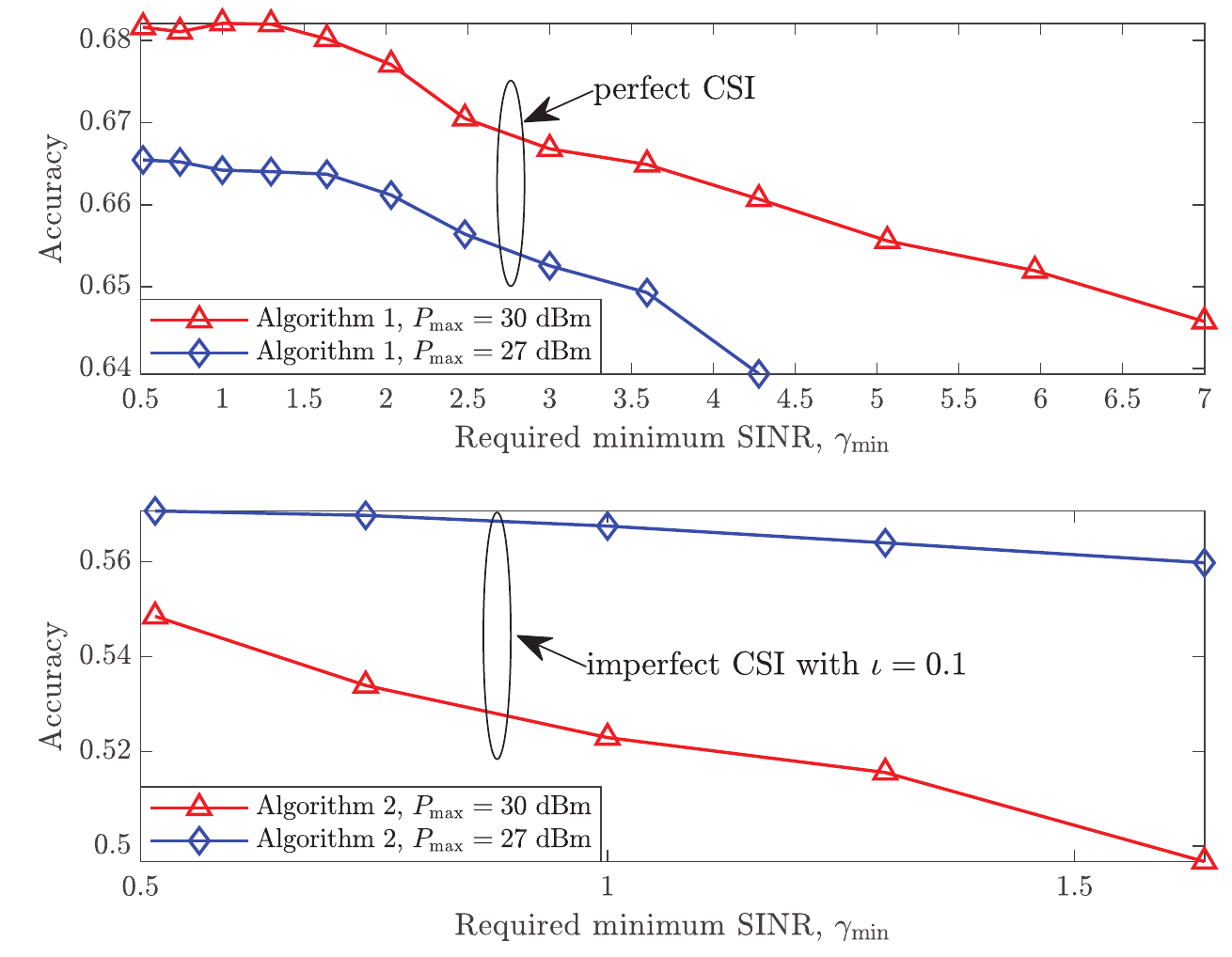}}
			\caption{Classification accuracy versus the required minimum SINR, where there are $2000$ training images per device, $K=3$, $N_t=2$, $N_r=16$, and $M=40$.}
			\label{accuracy_vs_R_min}
		\end{minipage}
		\vspace{-0.4 cm}
	\end{figure}

	To the best of our knowledge, no existing studies have captured the integrity of AirFL. 
	Let alone both the accuracy and integrity, as discussed in Section~\ref{contributions_and_organization}. 
	For this reason, no existing studies are directly comparable with the proposed Algorithms 1 and 2. 
	For comparison purpose, we plot the ideal FL, where the devices upload their local models separately and free of errors, and the BS aggregates the error-free local models to produce the global model which is then returned to the devices for continuing training. 
	The ideal FL can provide the upper bound for the classification accuracy of AirFL.
	
	Fig.~\ref{accuracy_vs_rounds} plots the classification accuracy of AirFL on the MNIST and Fashion-MNIST datasets with the growing number of communication rounds.
	We see that the classification accuracies of the proposed Algorithms~\ref{algorithm_1} and~\ref{algorithm_2} improve over rounds.
	Under perfect CSI (i.e., $\iota=0$), the accuracy of Algorithm 1 approaches the ideal FL, validating the effectiveness of Algorithm 1.
	As the result of channel estimation errors, the accuracy of Algorithm 2 decreases with the growth of $\iota$.
	In other words, the robustness of AirFL is at a cost of the classification accuracy.
	
	Fig.~\ref{accuracy_vs_R_min} shows the trade-off between the classification accuracy and the minimum SINR required for model recovery, under both perfect and imperfect CSI.
	On both the MNIST and Fashion-MNIST datasets, we see that the accuracies of the proposed algorithms slightly decline as the required minimum SINR increases, trading the accuracy for model integrity. 
	Under perfect CSI, increasing the transmit powers helps improve the trade-off by pushing the curves towards the top-right corner.
	In contrast, the increased transmit power can be detrimental under imperfect CSI, resulting from the interference caused by the channel estimation errors.
	
\begin{figure}
		\centering
		\begin{minipage}[t]{0.5 \textwidth}
			\centering
			\includegraphics[width=3.5 in]{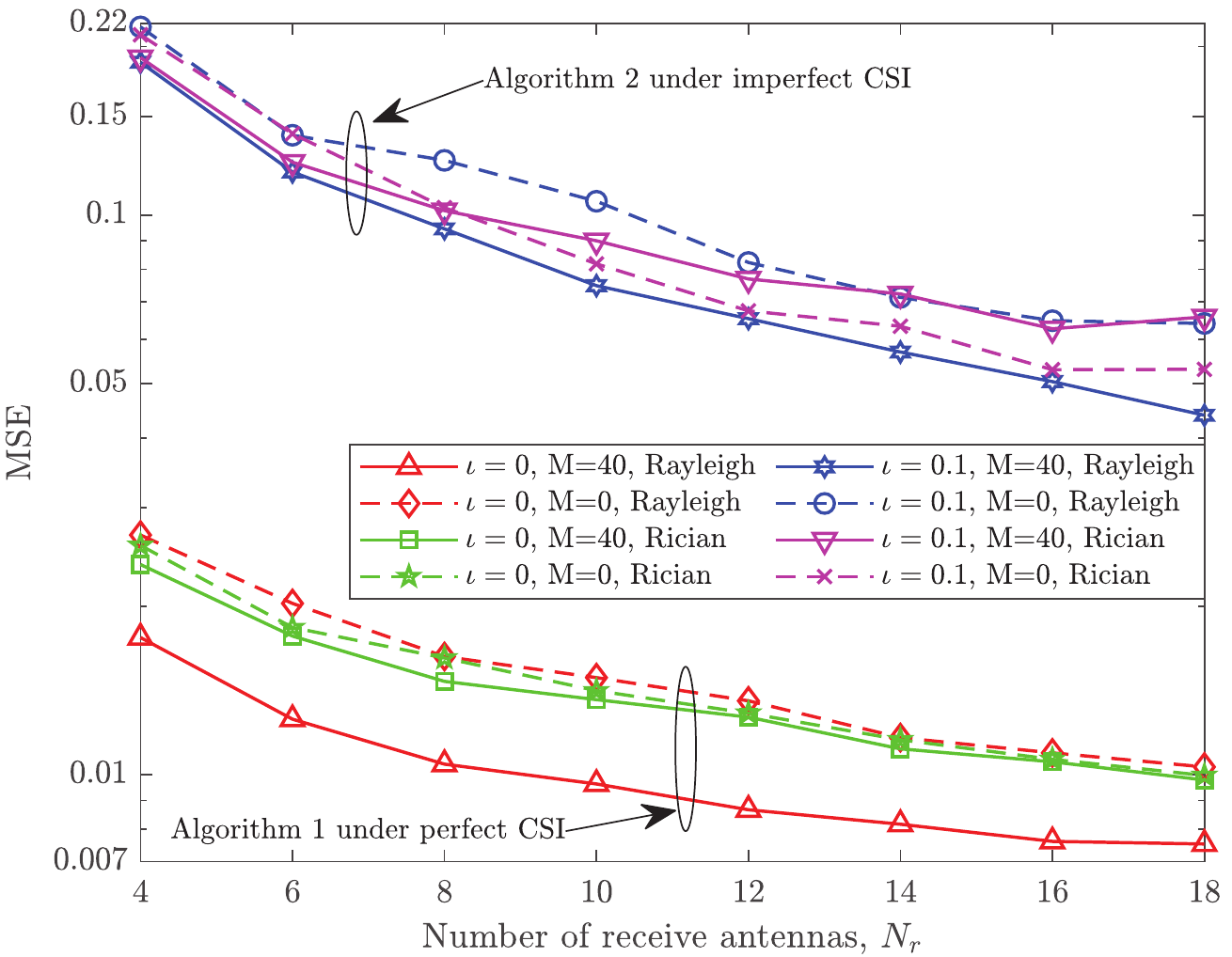}
			\caption{MSE versus the number of antennas, where ${\gamma_{\min}}=26.17$ dBm, $K=3$, and $N_t=2$.}
			\label{MSE_vs_antennas}
		\end{minipage}
	\end{figure}

	Fig.~\ref{MSE_vs_antennas} plots the MSE of Algorithms 1 and 2 with the increase of receive antennas at the BS.
	Here, $M=0$ indicates the case with no RIS.
	We consider two fading types for the device-BS channels, i.e., the Rician fading and Rayleigh fading.
	We see that the MSE of the global model declines with the increase of receive antennas.
	Under perfect CSI, the gain from employing the RIS is prominent, despite the gain diminishes with the increase of receive antennas under the Rician fading.
	This is because a large number of receive antennas provides sufficient array gain, hence overshadowing the improvement brought by the RIS.
	When the Rayleigh fading is considered, the MSE improvement of deploying the RIS is more significant due to the absence of the LoS.
	Under imperfect CSI, the use of the RIS can slow down the decline of the MSE with the increase of receive antennas under the Rician fading, and get outperformed by not using the RIS when the number of receive antennas is large, i.e., $N_r \ge 10$.
	This is due to severer interference caused by the channel estimation errors in the presence of more receive antennas.
	However, when the Rayleigh channel is considered, Algorithm 2 with the RIS can outperform the other considered scenarios. This confirms that the RIS is specifically desirable to improve the MSE when the LoS between the devices and the BS is blocked under imperfect CSI.
	
	\begin{figure}[!t]
		\centering
		\begin{minipage}[t]{0.5 \textwidth}
			\centering
			\subfigure[Accuracy on MNIST dataset versus $N_r$ with Rician fading between devices and the BS.]{
					\includegraphics[width=3.5 in]{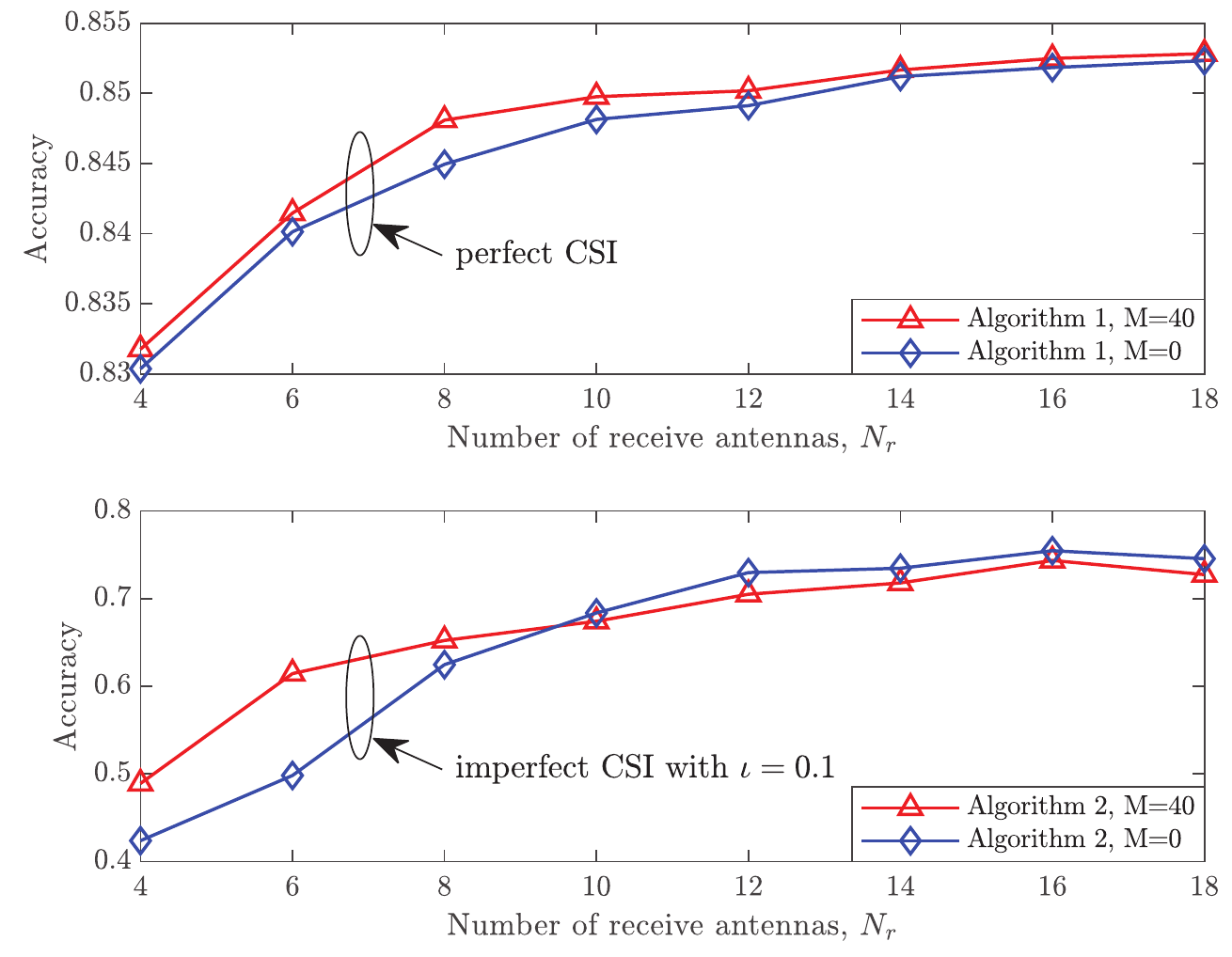}} \\
			\subfigure[Accuracy on  Fashion-MNIST dataset versus $N_r$ with Rayleigh fading between devices and the BS.]{
					\includegraphics[width=3.5 in]{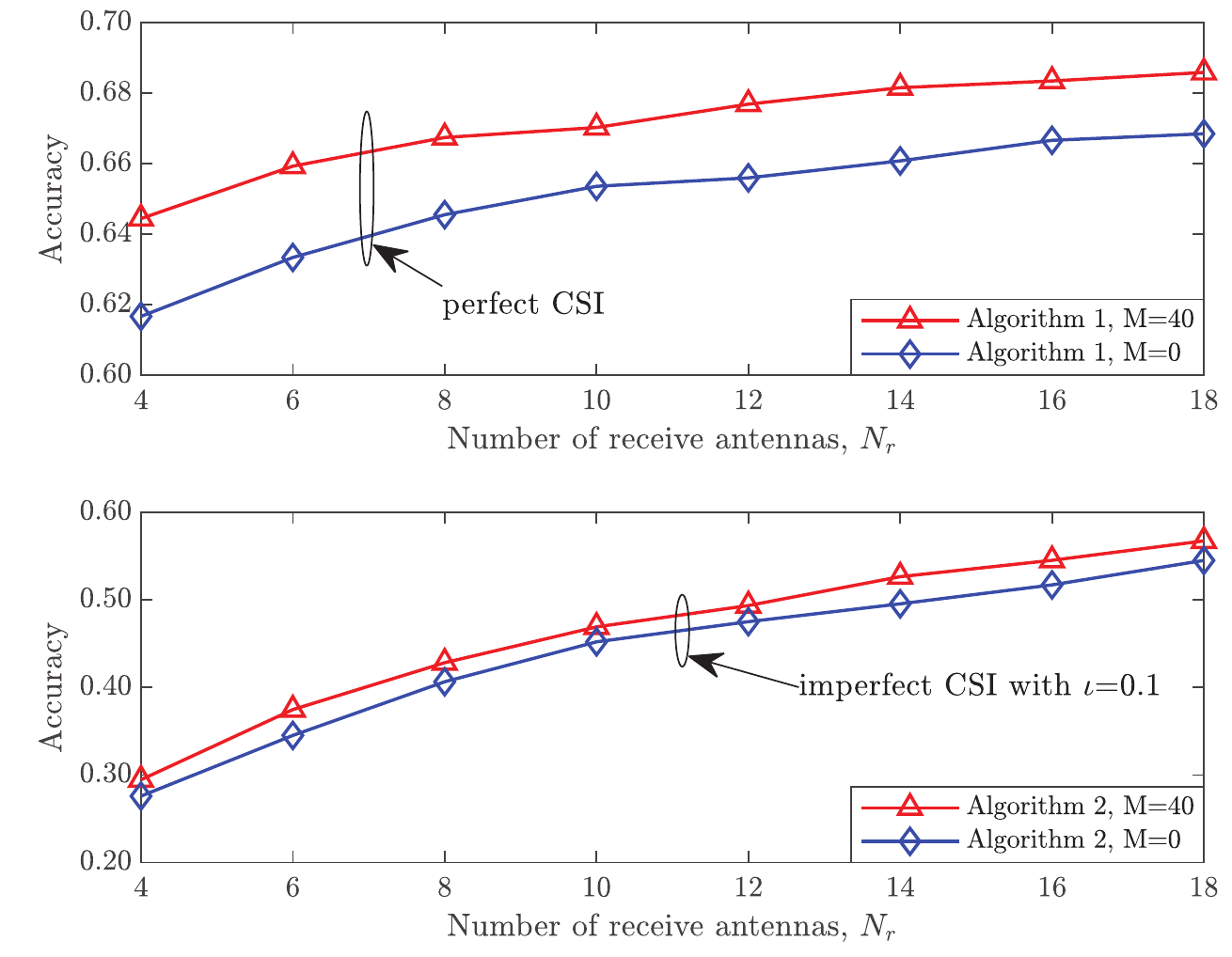}}
			\caption{Accuracy versus the number of antennas, where there are $2000$ training images per device, ${\gamma_{\min}}=26.17$ dBm, $K=3$, and $N_t=2$.}
			\label{accuracy_vs_antennas}
		\end{minipage}
	\end{figure}
	
	Fig.~\ref{accuracy_vs_antennas} shows the classification accuracy of AirFL on the MNIST and Fashion-MNIST datasets versus the number of receive antennas.
	In Fig.~\ref{accuracy_vs_antennas}(a), we see that the accuracy improves with the increase of receive antennas, resulting from the growing array gain.
	Under perfect CSI, the RIS clearly contributes to the improvement of the accuracy. 
	However, the contribution decreases with the increase of receive antennas under the Rician fading, because the array gain of the BS increasingly dominates.
	Under imperfect CSI (i.e., $\iota=0.1$), the RIS can help substantially improve the accuracy under the Rician fading when the number of receive antennas is small or moderate at the BS, e.g., $N_r\leq 8$. 
	The use of the RIS can result in a slightly reduced accuracy, when the number of receive antennas is large, e.g., $N_r\geq 10$. 
	This is because of the increased interference resulting from the estimation errors of the RIS-reflected channels.
	The conclusion drawn is that the RIS is beneficial for the accuracy of AirFL under the Rician fading and imperfect CSI, when the number of receive antennas is small or moderate.
	Nevertheless, we also see that deploying the RIS achieves higher accuracies under the Rayleigh fading in Fig.~\ref{accuracy_vs_antennas}(b).
	
\begin{figure}[t!]
		\centering
		\begin{minipage}[t]{0.5 \textwidth}
			\centering
			\includegraphics[width=3.2 in]{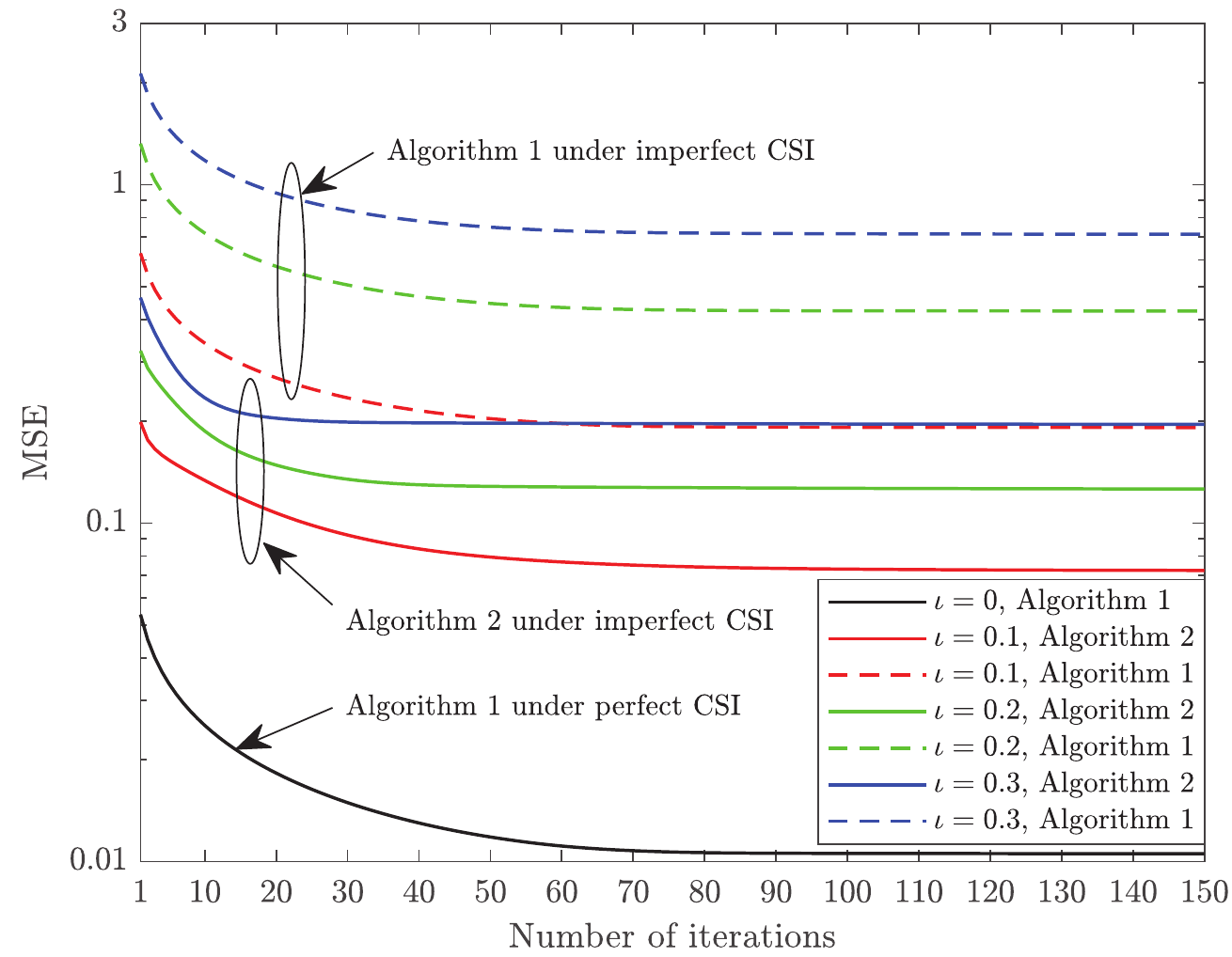}
			\caption{MSE vs the number of iterations, where ${\gamma_{\min}}=26.17$ dBm, $K=3$, $N_t=2$, $N_r=16$, and $M=40$.}
			\label{MSE_vs_iteration}
		\end{minipage}
	\end{figure}

	Fig.~\ref{MSE_vs_iteration} shows the convergence behaviors of our algorithms.
	We see that the MSE of both algorithms decreases monotonically over iterations until converge.
	Under perfect CSI (i.e., $\iota=0$), Algorithm 1 converges to the lowest MSE.
	Under imperfect CSI (i.e., $\iota>0$), the convergent MSE of the proposed algorithms grows with $\iota$.
	Under imperfect CSI, we also run Algorithm 1 designed for perfect CSI, and show that Algorithm 2 can substantially outperform Algorithm 1.
	This demonstrates the importance of the robust design under imperfect CSI.
	
	\begin{figure}
		\begin{minipage}[t]{0.5 \textwidth}
			\centering
			\includegraphics[width=3.2 in]{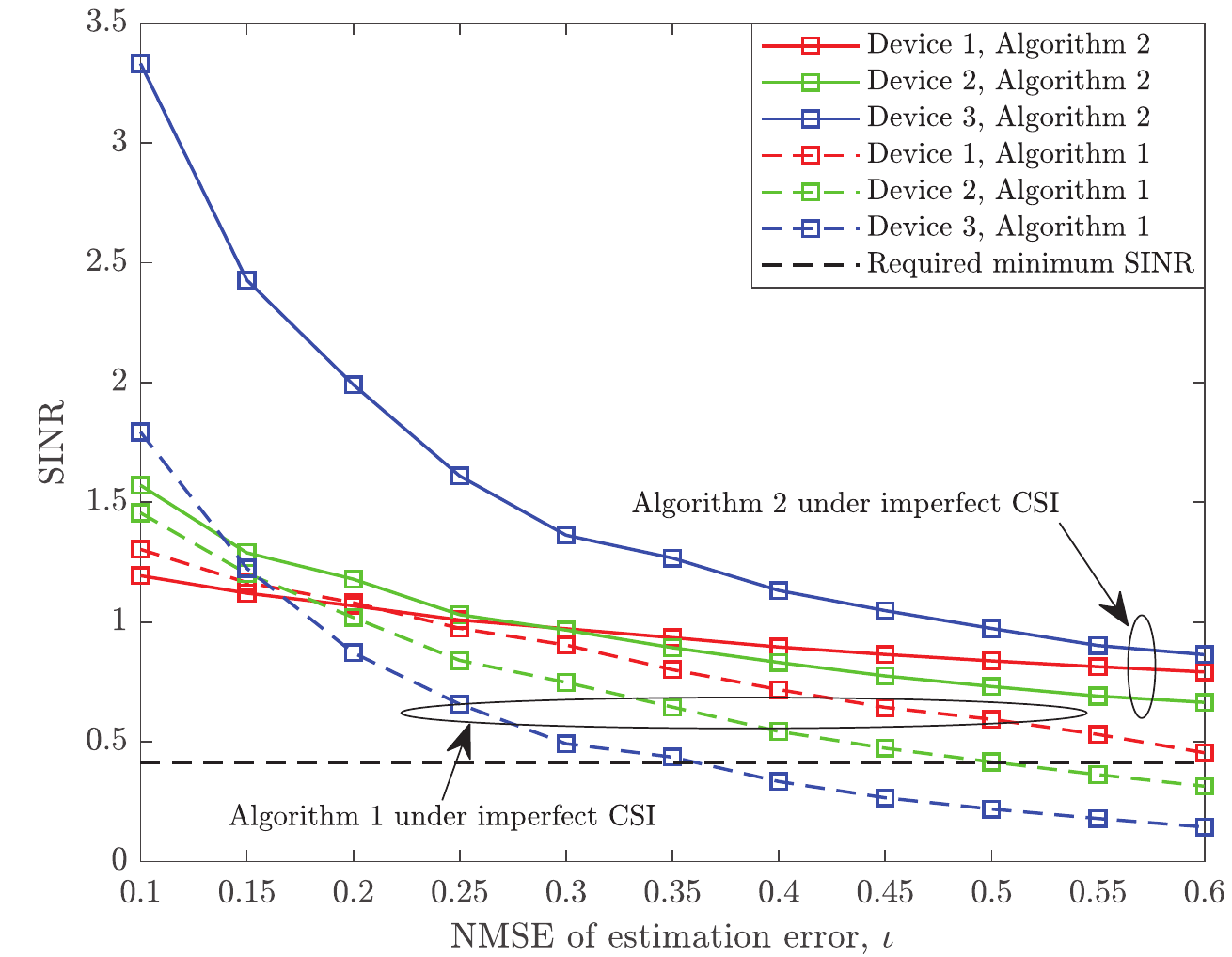}
			\caption{SINR versus the NMSE of channel estimation error, where ${\gamma_{\min}}=26.17$ dBm, $K=3$, $N_t=2$, $N_r=16$, and $M=40$.}
			\label{data_rate_vs_NMSE}
		\end{minipage}
		\vspace{-0.4 cm}
	\end{figure}
	
	Fig.~\ref{data_rate_vs_NMSE} plots the SINRs of Algorithms 1 and 2, versus the NMSE of channel estimation error $\iota$, under imperfect CSI.
	With the increase of $\iota$, the SINRs decline.
	Algorithm 2 achieves higher SINRs than the required minimum SINR $\gamma_{\rm min}$.
	This validates the robustness of Algorithm 2.
	In contrast, Algorithm 1 designed for perfect CSI may fail to guarantee the required minimum SINR (or in other words, the recoverability of the local models), when the channel estimation errors are non-negligible under imperfect CSI.
	
	\section{Conclusion} \label{conclusion}
	In this paper, we proposed a new framework to balance the accuracy and integrity for AirFL by designing two receive beamformers at the BS for AirFL and local model recovery.
	Under perfect CSI, we minimized the distortion of the aggregated model and retained the recoverability of the local models by optimizing the transmit and receive beamformers, and the RIS configuration in an alternating manner.
	Under imperfect CSI, we extended the framework to deliver a robust design of the beamformers and RIS configuration.
	Experiments showed that the framework achieves comparable learning accuracy and convergence to the ideal FL while preserving the local model recoverability under perfect CSI. Our framework also improves the accuracy when the number of receive antennas is small or moderate under imperfect CSI.

	\appendices
	\section{Transformation of  Problem (\ref{p_10})} \label{proof_of_proposition_1}
	Given fixed $\bf{b}$, $\{{\bf{a}}_k\}$ and $\bf{f}$, the objective (\ref{p_1_MSE}) is a function of $\bf{\Theta}$, denoted by $f_0({\bf{\Theta}})$. 
	Recalling the notations in Table~\ref{table_I}, we rewrite $f_0({\bf{\Theta}})$ as a function of $\bf{v}$, i.e., ${h_0}( {\bf{v}} )$:
	%
	%
	\begin{align}
		\label{MSE_theta_proof}
		f_0({\bf{\Theta}}) \overset{\quad}{=}& \sum\nolimits_{k=1}^K { {\bf{b}}^{\rm H}{ {\bf{G}}^{\rm H}}{\bf{\Theta}}{ {\bf{H}}}_{r,k}{\widehat {\bf{A}}}_k{ {\bf{H}}}^{\rm H}_{r,k}\bf{\Theta}}^{\rm H}{ {\bf{G}}{\bf{b}}} \notag \\
		&+ 2{\mathop{\rm Re}\nolimits} \{ ( {\bf{a}}^{\rm H}_k{ {\bf{H}}}^{\rm H}_{d,k}{\bf{b}}-1 ){\bf{b}}^{\rm H}{ {\bf{G}}}^{\rm H}{\bf{\Theta}}{ {\bf{H}}}_{r,k}{{\bf{a}}_k} \} \notag \\
		\overset{\quad}{=}& \text{tr}({{\bf{\Theta}}}^{\rm H}{{{\bf{\bar G}}}_b}{\bf{\Theta}}(\sum\limits_{k = 1}^K {{{\bf{Q}}_{0,k}}})) + 2{\mathop{\rm Re}\nolimits} \{ \text{tr}({(\sum\limits_{k = 1}^K {{{\bf{Q}}_{1,k}}})^{\rm H}}{\bf{\Theta}}) \} \notag \\
		\overset{(a)}{=}&{( {{e^{j{\bf{v}}}}} )^{\rm{H}}}{{\bf{F}}_0}{e^{j{\bf{v}}}}+2{\mathop{\rm Re}\nolimits} \{ {{( {{e^{j{\bf{v}}}}} )^{\rm H}}}{{{\bf{r}}_0}} \} \overset{\quad}{=} {h_0}( {\bf{v}} ),
	\end{align}
	where ${\widehat {\bf{A}}}_k = {\bf{a}}_k{\bf{a}}^{\rm H}_k, \forall k \in \mathcal{K}$.
	$(a)$ holds because $\text{tr}({\bf{A}}{\bf{\Theta}}{\bf{B}}{{\bf{\Theta}}}^{\rm H})=( {{e^{j{\bf{v}}}}} )^{\rm H}({\bf{A}} \circ {\bf{B}}^{\rm T}){{e^{j{\bf{v}}}}}$ and $\text{tr}({\bf{A}}^{\rm H}{\bf{\Theta}})=( {{e^{j{\bf{v}}}}} )^{\rm H}{{\rm vec}({\bf{A}})}$ for matrices ${\bf{A}}$ and ${\bf{B}}$, where ${{\rm vec}({\bf{A}})}$ vectorizes the diagonal of ${\bf{A}}$~\cite{Zeng2020joint}.
	%
	We write constraint (\ref{QoS_constraints_perfect_CSI}) as $f_{1,k}({\bf{\Theta}}) \le 0$ with $f_{1,k}({\bf{\Theta}})$ rewritten as ${h_{1,k}}( {\bf{v}} )$:
	%
	\begin{align}
		\label{constraint_1_proof}
		f_{1,k}({\bf{\Theta}}) \!\!=\!&-\!{\bf{f}}^{\rm H}{ {\bf{H}}}_k{\widehat {\bf{A}}}_k{ {\bf{H}}}^{\rm H}_k{\bf{f}} \!+\! {\gamma _{\min}} \sum\nolimits_{k'=k+1}^K \! {\bf{f}}^{\rm H}{ {\bf{H}}}_{k'}{\widehat {\bf{A}}}_{k'}{ {\bf{H}}}^{\rm H}_{k'}{\bf{f}} \notag \\
		&+ \! {\gamma _{\min}}{\sigma^2_n}{\|{\bf{f}}\|^2} \notag \\
		=&- \text{tr}({\bf{\Theta}}^{\rm H}{ {\bf{G}}}{\bf{f}}{\bf{f}}^{\rm H}{ {\bf{G}}^{\rm H}}{\bf{\Theta}}{ {\bf{H}}}_{r,k}{\widehat {\bf{A}}}_k{ {\bf{H}}}^{\rm H}_{r,k}) \notag \\
		&- \! 2{\mathop{\rm Re}\nolimits} \{ \text{tr}( { {\bf{H}}}_{r,k}{\widehat {\bf{A}}}_k{ {\bf{H}}}^{\rm H}_{d,k}{\bf{f}}{\bf{f}}^{\rm H}{ {\bf{G}}^{\rm H}}{\bf{\Theta}} ) \} + {C_{1,k}} \notag \\
		&+\!{\gamma _{\min}} \! \sum\nolimits_{k'=k+1}^K \! \{ \text{tr}(\!{\bf{\Theta}}^{\rm H}{ {\bf{G}}}{\bf{f}}{\bf{f}}^{\rm H}{ {\bf{G}}^{\rm H}}{\bf{\Theta}}{ {\bf{H}}}_{r,k'}{\widehat {\bf{A}}}_{k'}{ {\bf{H}}}^{\rm H}_{r,k'}\!) \notag \\ 
		&+ \! 2{\mathop{\rm Re}\nolimits} \{ \text{tr}( { {\bf{H}}}_{r,k'}{\widehat {\bf{A}}}_{k'}{ {\bf{H}}}^{\rm H}_{d,k'}{\bf{f}}{\bf{f}}^{\rm H}{ {\bf{G}}^{\rm H}}{\bf{\Theta}} )\} \} \notag \\
		=& -\text{tr}( {\bf{\Theta}}^{\rm H}{{{\bf{\bar G}}}_f}{\bf{\Theta}}( {{\bf{Q}}_{0,k}} \! - \! {\gamma _{\min}}\sum\nolimits_{k'=k+1}^K {{\bf{Q}}_{0,k'}} ) ) \notag \\ 
		&-\! 2{\mathop{\rm Re}\nolimits} \{ \text{tr}( ( {{\bf{Q}}^{\rm H}_{2,k}} \! - \! {\gamma _{\min}}\sum\nolimits_{k'=k+1}^K \! {{\bf{Q}}^{\rm H}_{2,k'}} ) {\bf{\Theta}} ) \} \! + \! {C_{1,k}} \notag \\
		=&- \! {( {{e^{j{\bf{v}}}}} )^{\rm{H}}}{{\bf{F}}_{1,k}}{e^{j{\bf{v}}}} - 2{\mathop{\rm Re}\nolimits} \{ {{{( {{e^{j{\bf{v}}}}} )}^{\rm{H}}}{{\bf{r}}_{1,k}}} \} + {C_{1,k}} \notag \\ =& {h_{1,k}}( {\bf{v}} ), \forall k \in \mathcal{K}.
	\end{align}
	Similarly, we write constraint (\ref{perfect_SIC_power_difference}) as $f_{2,k}({\bf{\Theta}}) \le 0$ with $f_{2,k}({\bf{\Theta}})$ rewritten as ${h_{2,k}}( {\bf{v}} )$:
	%
	%
	\begin{align}
		\label{constraint_2_proof}
		f_{2,k}({\bf{\Theta}})\!\! =& \!-\! {\bf{f}}^{\rm H}{ {\bf{H}}}_k{\widehat {\bf{A}}}_k{ {\bf{H}}}^{\rm H}_k{\bf{f}} \! + \! \sum\nolimits_{k'=k+1}^K \! {\bf{f}}^{\rm H}{ {\bf{H}}}_{k'}{\widehat {\bf{A}}}_{k'}{ {\bf{H}}}^{\rm H}_{k'}{\bf{f}} \!+\! {{\hat p}_{\rm gap}} \notag \\
		=& \!-\! {( {{e^{j{\bf{v}}}}} )^{\rm{H}}}{{\bf{F}}_{2,k}}{e^{j{\bf{v}}}} \! - \! 2{\mathop{\rm Re}\nolimits} \{ {{{( {{e^{j{\bf{v}}}}} )}^{\rm{H}}}{{\bf{r}}_{2,k}}} \} + C_{2,k} \notag \\ 
		=&{h_{2,k}}( {\bf{v}} ), \forall k \in \mathcal{K}\backslash\{K\}.
	\end{align}
	
	\vspace{-0.2 cm}
	\noindent As a result, problem (\ref{p_10}) is obtained.
	%
	
	\section{Proof of Lemma \ref{lemma_1}} \label{proof_of_lemma_1}
	The goal is to find constants $\{\xi _l\}$ that satisfy ${\xi _l}{{\bf{I}}_M} - {\nabla ^2}{h_l}\left( {\bf{v}} \right) \succeq {\bf{0}}, \forall l \in \mathcal{L}$.
	We take ${h_0}\left( {\bf{v}} \right)$ for example.
	${h_l}\left( {\bf{v}} \right), \forall l \in \mathcal{L}$ can be proved in the same way.
	The Hessian matrix of ${h_0}\left( {\bf{v}} \right)$, i.e., ${\nabla ^2}{h_0}\left( {\bf{v}} \right)$, can be decomposed into three parts, i.e., ${\nabla ^2}{h_0}\left( {\bf{v}} \right) = {\bf{P}}_1 + {\bf{P}}_2\left( {\bf{v}} \right) + {\bf{P}}_3\left( {\bf{v}} \right)$.
	%
	%
	\begin{align}
		\label{matrix_function_P_1}
		&{\bf{P}}_1 \buildrel \Delta \over = {\rm diag}( 2{{{[{{\bf{F}}_0}]}_{1,1}}},\cdots,2{{{[{{\bf{F}}_0}]}_{M,M}}} ), \\
		\label{matrix_function_P_2}
		&{\bf{P}}_2\left( {\bf{v}} \right) \buildrel \Delta \over = {\rm diag}( {\!-\! 2{\mathop{\rm Re}\nolimits} \{ {{e^{j{\phi _1}}}( {\sum\nolimits_{i = 1}^M {{e^{ - j{\phi _i}}}{{[{{\bf{F}}_0}]}_{i,1}}} }\! +\! [{{\bf{r}}_0}]_1^* )} \}} ,\cdots, \notag \\ 
		&\quad\quad\quad\quad{\!-\! 2{\mathop{\rm Re}\nolimits} \{ {{e^{j{\phi _M}}}( {\sum\nolimits_{i = 1}^M {{e^{ - j{\phi _i}}}{{[{{\bf{F}}_0}]}_{i,M}}} }\! +\! [{{\bf{r}}_0}]_M^* )} \}} ),
	\end{align}
	\begin{figure*}
	\begin{align}
			\label{matrix_function_P_3}
			{{\bf{P}}_3}\left( {\bf{v}} \right) \buildrel \Delta \over =& \left( {\begin{array}{*{20}{c}}
					0&{2{\rm{Re\{ }}{e^{j\left( {{\phi _1} - {\phi _2}} \right)}}{{[{{\bf{F}}_0}]}_{2,1}}{\rm{\} }}}& \cdots &{2{\rm{Re\{ }}{e^{j\left( {{\phi _1} - {\phi _M}} \right)}}{{[{{\bf{F}}_0}]}_{M,1}}{\rm{\} }}}\\
					{2{\rm{Re\{ }}{e^{j\left( {{\phi _2} - {\phi _1}} \right)}}{{[{{\bf{F}}_0}]}_{1,2}}{\rm{\} }}}&0& \cdots &{2{\rm{Re\{ }}{e^{j\left( {{\phi _2} - {\phi _M}} \right)}}{{[{{\bf{F}}_0}]}_{M,2}}{\rm{\} }}}\\
					\vdots & \vdots & \ddots & \vdots \\
					{2{\rm{Re\{ }}{e^{j\left( {{\phi _M} - {\phi _1}} \right)}}{{[{{\bf{F}}_0}]}_{1,M}}{\rm{\} }}}&{2{\rm{Re\{ }}{e^{j\left( {{\phi _M} - {\phi _2}} \right)}}{{[{{\bf{F}}_0}]}_{2,M}}{\rm{\} }}}& \cdots &0
			\end{array}} \right).
		\end{align}
	\hrulefill
	\end{figure*}
	
	\vspace{-0.4 cm}
	\noindent Likewise, we rewrite ${\xi _0}{\bf{I}}_M$ as ${\xi _0}{\bf{I}}_M={\xi _{0,1}}{\bf{I}}_M + {\xi _{0,2}}{\bf{I}}_M + {\xi _{0,3}}{\bf{I}}_M$.
	Then, ${\xi _0}{{\bf{I}}_M} - {\nabla ^2}{h_0}\left( {\bf{v}} \right) \succeq {\bf{0}}$ is replaced by three more stringent constraints ${\xi _{0,i}}{\bf{I}}_M-{\bf{P}}_i\left( {\bf{v}} \right) \succeq {\bf{0}}, \forall i \in \{1,2,3\}$.

	As for ${\xi _{0,1}}$, we have ${\xi _{0,1}}{\bf{I}}_M-{\bf{P}}_1 \succeq {\bf{0}}$ if ${\xi _{0,1}} \ge 2\mathop {\max }\limits_{m \in {\cal M}} \{ {| {{{[ {{{\bf{F}}_0}} ]}_{m,m}}} |} \}$.
	
	As for ${\xi _{0,2}}$, we have the following inequality:
	\vspace{-0.2 cm}
	\begin{align}
		\overset{}{-} & 2{\mathop{\rm Re}\nolimits}  \{ {{e^{j{\phi _m}}}( {\sum\nolimits_{i = 1}^M {{e^{ - j{\phi _i}}}{{[{{\bf{F}}_0}]}_{i,m}}} }\! +\! [{{\bf{r}}_0}]_m^* )} \}  \notag \\
		\overset{}{=} & 2 (\sum\nolimits_{i = 1}^M {|{[{{\bf{F}}_0}]}_{i,m}|{\rm cos}({\phi _m}-{\phi _i}+{\phi _{{[{{\bf{F}}_0}]}_{i,m}}}+\pi)} \notag \\
		&+ |[{{\bf{r}}_0}]_m|{\rm cos}({\phi _m}-{\phi _{[{{\bf{r}}_0}]_m}} \pm \pi) )  \notag \\
		\overset{(a)}{\le} & 2( {\sum\nolimits_{i = 1}^M {|{{[{{\bf{F}}_0}]}_{i,m}}|}  + |{{[{{\bf{r}}_0}]}_m}|} ), \forall m \in \mathcal{M},
	\end{align}

	\noindent where $[{{\bf{r}}_0}]_m^*$ denotes the conjugate of $[{{\bf{r}}_0}]_m$, and $(a)$ stems from ${\rm cos}(x) \le 1, \forall x \in [0,2\pi)$.
	Therefore, we have ${\xi _{0,2}}{\bf{I}}_M-{\bf{P}}_2\left( {\bf{v}} \right) \succeq {\bf{0}}$ if ${\xi _{0,2}} \ge 2\mathop {\max }\limits_{m \in {\cal M}} \{ {\sum\nolimits_{i = 1}^M {| {{{[ {{{\bf{F}}_0}} ]}_{i,m}}} |}  + | {{{[ {{{\bf{r}}_0}} ]}_m}} |} \}$.
	
	As for ${\xi _{0,3}}$, we find that ${{\bf{P}}_3}\left( {\bf{v}} \right)$ can be decomposed as
	%
	%
	\begin{align}
		{{\bf{P}}_3}\left( {\bf{v}} \right) \overset{\quad}{=}& 2{\mathop{\rm Re}\nolimits} \{ {\bf{\Theta}} {{\bar{\bf{F}}}_0} {\bf{\Theta}}^{\rm H} \} \notag \\
		\overset{(a)}{=} & 2{\mathop{\rm Re}\nolimits} \{ {\bf{\Theta}} {\bf{U}} {\bf{\Lambda}} {\bf{U}}^{\rm H} {\bf{\Theta}}^{\rm H} \} \notag \\
		\overset{(b)}{=} & {\widetilde {\bf{U}}}({\bf{v}}) (2{\bf{\Lambda}}) ({\widetilde {\bf{U}}}({\bf{v}}))^{\rm H},
	\end{align}

	\noindent where $\bar{\bf{F}}_0=(\Psi({\bf{F}}_0))^{\rm T}$ with $\Psi({\bf{F}}_0)$ setting all elements along the main diagonal of ${\bf{F}}_0$ to zero.
	$(a)$ is obtained by performing singular value decomposition of ${{\bar{\bf{F}}}_0}$, i.e., ${{\bar{\bf{F}}}_0} = {\bf{U}} {\bf{\Lambda}} {\bf{U}}^{\rm H}$, with ${\bf{\Lambda}}$ being a diagonal matrix, and ${\bf{U}}$ being a unitary matrix.
	$(b)$ is due to the fact that the singular values of the Hermitian matrix ${{\bar{\bf{F}}}_0}$ are real numbers, and ${\widetilde {\bf{U}}}({\bf{v}}) \buildrel \Delta \over = {\bf{\Theta}} {\bf{U}}$.

	We note that the singular values of the Hermitian matrix ${{\bar{\bf{F}}}_0}$ are intrinsically non-negative~\cite{Golub2013Matrix}. Therefore, the maximum singular value of ${{\bar{\bf{F}}}_0}$ (which is also the maximum element on the main diagonal of ${\bf{\Lambda}}$), denoted by $\omega _{{{\bar{\bf{F}}}_0}}$, is non-negative, i.e., ${\omega _{{{\bar{\bf{F}}}_0}}}\ge0$.
	If ${\xi _{0,3}} \ge 2{\omega _{{{\bar{\bf{F}}}_0}}}$, then
	%
	%
	\begin{align}
		{\bf{x}}^{\rm H}[{\xi _{0,3}}{\bf{I}}_M \! - \!{{\bf{P}}_3}\left( {\bf{v}} \right)]{\bf{x}} \ge & {\bf{x}}^{\rm H}{\widetilde {\bf{U}}}({\bf{v}})(2\omega _{{{\bar{\bf{F}}}_0}}{\bf{I}}_M \! -\! 2{\bf{\Lambda}})({\bf{x}}^{\rm H}{\widetilde {\bf{U}}}({\bf{v}}))^{\rm H} \notag \\ 
		\ge & 0, \forall {\bf{x}} \in {\mathbb{C}}^{M \times 1}.
	\end{align}

	\noindent Clearly, ${\xi _{0,3}}{\bf{I}}_M-{{\bf{P}}_3}\left( {\bf{v}} \right) \succeq {\bf{0}}$.

	As a result, we have ${\xi _0}{{\bf{I}}_M} \succeq {\nabla ^2}{h_0}\left( {\bf{v}} \right)$, if
	\begin{align}
		{\xi _0}=&{\xi _{0,1}}+{\xi _{0,2}}+{\xi _{0,3}} \notag \\ 
		\ge& 2\mathop {\max }\limits_{m \in {\cal M}} \left\{ {\sum\nolimits_{i = 1}^M {| {{{[ {{{\bf{F}}_0}} ]}_{i,m}}} |}  + | {{{[ {{{\bf{r}}_0}} ]}_m}} |} \right\} + 2{\omega _{{{\bar{\bf{F}}}_0}}} \notag \\
		&+ 2\mathop {\max }\limits_{m \in {\cal M}} \left\{ {| {{{[ {{{\bf{F}}_0}} ]}_{m,m}}} |} \right\}.
	\end{align}
	
	\section{Proof of Lemma \ref{lemma_2}} \label{proof_of_lemma_2}
	For the desired aggregated model $s$ and the superposition signal ${{\tilde s}_b}$, the MSE is given by
	\vspace{-0.2 cm}
		\begin{align}
			\label{MSE_imperfect_detailed}
			\hspace{-0.3 cm}
			{\rm MSE}\left( {\tilde s}_b,s \right)\!\! =&  \mathbb{E}[\left( {\tilde s}_b \!-\! s \right)^{\rm H}\left( {\tilde s}_b \!-\! s \right)] \notag \\
			=& \! \underbrace{\mathbb{E}[ | {\bf{b}}^{\rm H}{\bf{n}} |^2 ]}_{{\rm MSE}_1} \! + \! \underbrace{\mathbb{E}[ | \sum\nolimits_{k = 1}^K { ( {\bf{b}}^{\rm H}\tilde {\bf{H}}_k {\bf{a}}_k \!- \! 1 ) s_k } |^2 ]}_{{\rm MSE}_2} \notag \\ 
			&+ \! \underbrace{2{\mathop{\rm Re}\nolimits} \{\mathbb{E}[ \sum\nolimits_{k = 1}^K \! { \! ({\bf{n}}^{\rm H}{\bf{b}}) ( {\bf{b}}^{\rm H}\tilde {\bf{H}}_k {\bf{a}}_k \!-\! 1 ) s_k }  ]}_{{\rm MSE}_3} \}, \!\!\!\!
	\end{align}
	where $\tilde {\bf{H}}_k = {\widehat {\bf{H}}}_k + {\Delta {\bf{H}}_{k}}, \forall k \in \mathcal{K}$.
	Then, ${\rm MSE}_1$ can be written as
	\begin{align}
		{\rm MSE}_1 =& \sum\nolimits_{i=1}^{N_r} {\left| \left[ {\bf{b}} \right]_i \right|}^2 \mathbb{E}[ {\left| \left[ {\bf{n}} \right]_i \right|}^2 ] \notag \\
		=& \sigma^2_n{\left\| {\bf{b}} \right\|^2}.
	\end{align}
	${\rm MSE}_2$ can be rewritten as
	\begin{align}
		{\rm MSE}_2 =& \sum\nolimits_{k=1}^K \! {{\left|{{\bf{b}}^{\rm H}}{\widehat {\bf{H}}_k}{{\bf{a}}_k} - 1\right|}^2} \notag \\
		& + \sum\nolimits_{k=1}^K \left\{ {\bf{b}}^{\rm H} ( \mathbb{E}[ {\Delta {\bf{H}}_{d,k}}{ {\widehat {\bf{A}}}_k }{\Delta {\bf{H}}^{\rm H}_{d,k}} ] \right. \notag \\ 
		&+ \mathbb{E}[{\widehat {\bf{G}}^{\rm H}}{\bf{\Theta}}{\Delta {\bf{H}}_{r,k}}{{\widehat {\bf{A}}}_k}{\Delta {\bf{H}}^{\rm H}_{r,k}}{\widehat {\bf{G}}} ] \notag \\
		&+ \mathbb{E}[{{\Delta {\bf{G}}}^{\rm H}}{{\bf{\Theta}}}{{\widehat {\bf{H}}_{r,k}}}{{\widehat {\bf{A}}}_k}{\widehat {\bf{H}}^{\rm H}_{r,k}}{{\bf{\Theta}}^{\rm H}}{\Delta {\bf{G}}}] \notag \\ 
		&\left. + \mathbb{E}[{{\Delta {\bf{G}}}^{\rm H}}{{\bf{\Theta}}}{{\Delta {\bf{H}}_{r,k}}}{{\widehat {\bf{A}}}_k}{\Delta {\bf{H}}^{\rm H}_{r,k}}{{\bf{\Theta}}^{\rm H}}{\Delta {\bf{G}}}]){\bf{b}} \right\}.
	\end{align}
	
	\vspace{-0.2 cm}
	We employ the conclusions of~\cite{Rong2011Robust} and~\cite{Gupta2000Matrix}: Given constant matrices $\bf{A}$, $\bf{B}$, $\bf{C}$ and a random matrix $\bf{X}$ yielding ${\bf{X}} \sim{} \mathcal{CN}\left( {\bar {\bf{X}}}, {{\bf{\Sigma }}} \otimes {{\bf{\Psi }}} \right)$, we have $\mathbb{E}[ {\bf{X}}{\bf{A}}{\bf{X}}^{\rm H} ] = {\bar {\bf{X}}}{\bf{A}}{\bar {\bf{X}}}^{\rm H} + \text{tr}({\bf{C}}{{\bf{\Sigma }}}^{\rm T}){{\bf{\Psi }}}$ and $\mathbb{E}[{\bf{A}}{\bf{X}}{\bf{B}}{\bf{X}}^{\rm H}{\bf{C}}]={\bf{A}}\mathbb{E}[{\bf{X}}{\bf{B}}{\bf{X}}^{\rm H}]{\bf{C}}$.
	Then, ${\rm MSE}_2$ can be further rewritten as
	%
	%
	\begin{align}
		{\rm MSE}_2 =&\!\sum\nolimits_{k=1}^K \! {{\left|{{\bf{b}}^{\rm H}}{\widehat {\bf{H}}_k}{{\bf{a}}_k} - 1\right|}^2} \notag \\ 
		&+ \sum\nolimits_{k=1}^K {\bf{b}}^{\rm H} \! \left[ {\sigma^2_{d,k}}\text{tr}({\widehat {\bf{A}}}_k){\bf{I}}_{N_r} + \! {\sigma^2_{r,k}}\text{tr}({\widehat {\bf{A}}}_k){{\widehat {\bf{G}}}^{\rm H}}{{\widehat {\bf{G}}}} \right. \notag \\
		&\left. + {\sigma^2_g}\text{tr}({{\widehat {\bf{H}}}_{r,k}}{{\widehat {\bf{A}}}_k}{{\widehat {\bf{H}}}^{\rm H}_{r,k}}){\bf{I}}_{N_r} + M{\sigma^2_{r,k}}{\sigma^2_g}\text{tr}({\widehat {\bf{A}}}_k){\bf{I}}_{N_r} \right]\!{\bf{b}} \notag \\
		=&\sum\nolimits_{k=1}^K \! {{\left|{{\bf{b}}^{\rm H}}{\widehat {\bf{H}}_k}{{\bf{a}}_k} - 1\right|}^2} + {\sum\nolimits_{k = 1}^K {{{\bf{b}}^{\rm H}}{{\bf{J}}_k}{\bf{b}}}},
	\end{align}
	
	\vspace{-0.2 cm}
	\noindent where ${\widehat {\bf{A}}}_k$ is given in Appendix \ref{proof_of_proposition_1}.
	${\rm MSE}_3$ can be written as
	%
	\begin{align}
		{\rm MSE}_3 = & 2{\mathop{\rm Re}\nolimits} \left\{ \! \sum\nolimits_{k = 1}^K  \! \mathbb{E} [ { ({\bf{n}}^{\rm H}{\bf{b}}) ( {\bf{b}}^{\rm H}\tilde {\bf{H}}_k {\bf{a}}_k ) s_k \! }  -  ({\bf{n}}^{\rm H}{\bf{b}}) s_k  ] \! \right\} \notag \\ 
		=& 0.
	\end{align}
	
	\vspace{-0.2 cm}
	As a result, we have
	%
	\begin{align}
		{\rm MSE}\!\left( {\tilde s}_b,\!s \right)\!=\!\! {\sum\nolimits_{k = 1}^K \! {|{{\bf{b}}^{\rm H}}{\widehat {\bf{H}}_k}{{\bf{a}}_k} \!-\!\! 1|}^2} \!\!+\! {{\sigma}^2_n}{\left\| {\bf{b}} \right\|^2} \!\!+\! {\sum\nolimits_{k = 1}^K \!\! {{{\bf{b}}^{\rm H}}{{\bf{J}}_k}{\bf{b}}}}.
	\end{align}
	
	\bibliographystyle{IEEEtran}
	\bibliography{IEEEabrv,ref}

\end{document}